\documentclass[acmsmall]{acmart}

\AtBeginDocument{%
  \providecommand\BibTeX{{%
    \normalfont B\kern-0.5em{\scshape i\kern-0.25em b}\kern-0.8em\TeX}}}

\setcopyright{acmcopyright}
\copyrightyear{2021}
\acmYear{2021}
\acmDOI{10.1145/1122445.1122456}

\acmJournal{TECS}
\acmVolume{37}
\acmNumber{4}
\acmArticle{111}
\acmMonth{8}

\usepackage[redeflists]{IEEEtrantools}
\DeclareMathAlphabet{\mymathbb}{U}{BOONDOX-ds}{m}{n}
{}
{}
{}
\newcommand{\ineq}[1]{\footnotesize$#1$\normalsize}{}
\usepackage[linesnumbered,ruled,noend]{algorithm2e}
\usepackage{threeparttable}
\usepackage{amsmath}
\newtheorem{Tlemma}{Lemma}

\newtheorem{Tdef}{Definition}

\usepackage{multirow}
\usepackage{xcolor}
\usepackage{subfig}

\SetCommentSty{mycommfont}    
\usepackage{pifont}
\let\oldding\ding
\renewcommand{\ding}[2][1]{\scalebox{#1}{\oldding{#2}}}
\newcommand{\mr}[1]{\textcolor{black}{#1}}
\newcommand{\minor}[1]{\textcolor{black}{#1}}

\newcommand{\ineqq}[1]{\scriptsize$#1$\normalsize}{}

\usepackage{tabu}
\usepackage{color, colortbl}
\newcommand{\thickhline}{%
    \noalign {\ifnum 0=`}\fi \hrule height 1pt
    \futurelet \reserved@a \@xhline
}

\newcommand{\mubrain}{$\mu$\text{{Brain}}}{}
\newcommand{\ckts}{\text{{circuitries}}}{}
\begin{document}
\bstctlcite{IEEEexample:BSTcontrol}

\title{Design-Technology Co-Optimization for NVM-based Neuromorphic Processing Elements}

\author{Shihao Song}
\author{Adarsha Balaji}
\author{Anup Das}
\author{Nagarajan Kandasamy}
\email{anup.das@drexel.edu}
\affiliation{%
  \institution{Drexel University}
  \streetaddress{3141 Chestnut Street}
  \city{Philadelphia}
  \state{PA}
  \country{USA}
  \postcode{19104}
}

\renewcommand{\shortauthors}{Song, et al.}

\begin{abstract}
    \mr{
An emerging use-case of machine learning (ML) is to train a model on a high-performance system and deploy the trained model on energy-constrained embedded systems. Neuromorphic hardware platforms, which operate on principles of the biological brain, can significantly lower the energy overhead of a machine learning inference task, making these platforms an attractive solution for embedded ML systems.
}
We present a design-technology tradeoff analysis to implement such inference tasks on the processing elements (PEs) of a Non-Volatile Memory (NVM)-based neuromorphic hardware. 
Through detailed circuit-level simulations at scaled process technology nodes, we show the negative impact of technology scaling on the information-processing latency, \mr{which impacts the quality-of-service (QoS) of an embedded ML system}. 
At a finer granularity, the latency inside a PE depends on 1) the delay introduced by parasitic components on its current paths, and 2) the varying delay to sense different resistance states of its NVM cells.
Based on these two observations, we make the following three contributions. First, on the technology front, we propose an optimization scheme where the NVM resistance state that takes the longest time to sense is set on current paths having the least delay, and vice versa, \mr{reducing the average PE latency, which improves the QoS.} 
Second, on the architecture front, we introduce isolation transistors within each PE to partition it into regions that can be individually power-gated, reducing both latency and energy.
Finally, on the system-software front, we propose a mechanism to leverage the proposed technological and architectural enhancements when implementing a machine-learning inference task on neuromorphic PEs of the hardware. 
Evaluations with a recent neuromorphic hardware architecture show that our proposed design-technology co-optimization approach improves both performance and energy efficiency of machine-learning inference tasks without incurring high cost-per-bit.
\end{abstract}



\maketitle

\section{Introduction}\label{sec:introduction}
Neuromorphic computing systems are integrated circuits that implement the architecture of central nervous system in primates~\cite{mead1990neuromorphic,bose2019my,christensen20212021}. 
\mr{
These systems facilitate energy-efficient computations using Spiking Neural Networks (SNN)~\cite{maass1997networks} for power-constrained embedded devices. To this end, the design workflow is to train a machine learning model (commonly on a backend server) and subsequently, convert the trained model to spike-based computations and deploy it on the neuromorphic hardware of an embedded system.
The quality of inference (e.g., accuracy) is assessed in terms of the inter-spike interval (ISI) (see Section~\ref{sec:isi_distortion}). Therefore, any deviation from its expected value will lead to a degradation of the inference quality.
}

\begin{figure}[h!]
	\centering
	\vspace{-10pt}
	\centerline{\includegraphics[width=0.99\columnwidth]{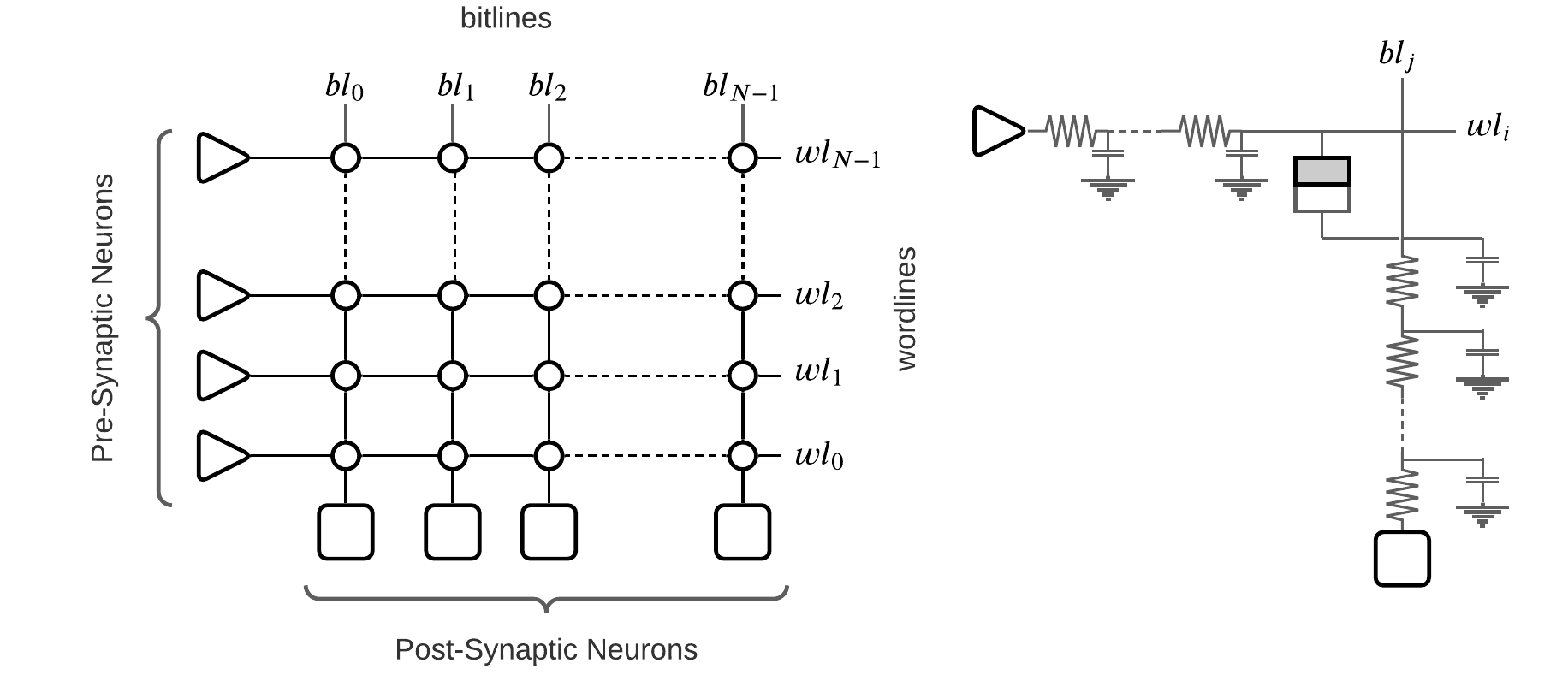}}
	\caption{An $N \times N$ crossbar showing the parasitic components within.}
	\vspace{-10pt}
	\label{fig:crossbar}
\end{figure}

\mr{
A typical neuromorphic system such as Loihi~\cite{loihi}, DYNAPs~\cite{dynapse} and \mubrain{}~\cite{sentryos} consists of processing elements (PEs) that communicate spikes using a shared interconnect. Each PE implements neuron and synapse \ckts{}. 
}
A common technique to implement a neuromorphic PE is using an analog crossbar where 
bitlines and wordlines are organized in a grid with memory cells connected at their crosspoints to store synaptic weights~\cite{liu2015spiking,hu2014memristor,hu2016dot,ankit2017trannsformer,nukala2014spintronic,kim2012digital,zhang2018neuromorphic,gopalakrishnan2020hfnet,fernando20203d}.
Neuron \ckts{} are implemented along bitlines and wordlines. 
\mr{
Figure~\ref{fig:crossbar} (left) shows the architecture of an \ineq{N\times N} analog crossbar with \ineq{N} bitlines and \ineq{N} wordlines. 
}

\mr{
We investigate the internal architecture of a crossbar and find that parasitic components introduce delay in propagating current from a pre-synaptic neuron to a  post-synaptic neuron as illustrated in Figure~\ref{fig:crossbar} (right). 
This delay depends on the specific current path activated in a crossbar.
Higher the number of parasitic components on a current path, larger is its propagation delay~\cite{paul2021design,twisha_endurance,twisha_thermal,espine,song2021improving}. Parasitic components on bitlines and wordlines are a major source of latency at scaled process technology nodes and they create significant \textbf{latency variation} in a crossbar~\cite{mneme,hebe,lee2013tiered,mutlu2013memory,thomas2021analysis,fouda2017modeling,jeong2017parasitic,krestinskaya2019memristive,fouda2018overcoming}. 
Such variation can introduce ISI distortion (Section~\ref{sec:isi_distortion}), which may impact the quality of an inference task~\cite{spinemap,psopart}.
}

To increase the energy efficiency of a neuromorphic system, Non-Volatile Memory (NVM) such as oxide-based random access memory (OxRRAM), phase-change memory (PCM), ferroelectric RAM (FeRAM), and spin-based magnetic RAM (MRAM) is used to implement the memory cells in a crossbar~\cite{mallik2017design,Burr2017,wijesinghe2018all,young2019evolving,wang2020ncpower}.
An NVM cell can be programmed to a high-resistance state (HRS) or one of many low-resistance states (LRS), implementing multi-bit synaptic weights~\cite{mallik2017design,xue201924,liao2021multibit}.
\mr{
To implement a synaptic weight on a memory cell of a crossbar, the synaptic weight is programmed as the conductance of the cell.
}

A crossbar can accommodate only a fixed number of pre-synaptic connections per post-synaptic neuron. 
\mr{
To give an an example, the crossbar in Figure~\ref{fig:crossbar} (left) has \ineq{N} pre-synaptic neurons, \ineq{N} post-synaptic neurons, and \ineq{N^2} memory cells. Each post-synaptic neuron can have a maximum of \ineq{N} pre-synaptic connections.
To mitigate the negative impact of technology scaling, e.g., increase in the value of parasitic components on current paths and increase in the power density, \ineq{N} is constrained to a lower value, typically between 128 and 256 (see our tradeoff analysis in Section~\ref{sec:tradeoff}).
}
To map a large SNN model on a multi-PE hardware, a system software framework such as NEUTRAMS~\cite{ji2016neutrams}, NeuroXplorer~\cite{neuroxplorer}, SentryOS~\cite{sentryos}, LAVA~\cite{loihi_mapping} and DFSynthsizer~\cite{dfsynthesizer_pp} is commonly used. 
\mr{
These frameworks first partition a model into clusters, where a cluster is a subset of neurons and synapses of the model that can be implemented on the architecture of a crossbar. Subsequently, the partitioned clusters are implemented on different crossbars of a neuromorphic hardware.
}

We make the following two key \textbf{observations} related to a neuromorphic PE.

\emph{\textbf{Observation 1:} \mr{ The latency within a crossbar is a function of the length (i.e., the number of parasitic components) of current paths and the delay to sense the NVM cell activated on a current path.}}

\emph{\textbf{Observation 2:} \mr{Due to how memory cells are organized in a crossbar, a significant fraction of these memory cells remains unutilized when implementing machine learning inference tasks.}}

Based on these two observations (which we elaborate in  Sections~\ref{sec:tradeoff}-\ref{sec:architecture_optimization}), we present a design-technology tradeoff analysis to implement machine learning inference tasks on different PEs of an NVM-based neuromorphic system.
We make the following four key \textbf{contributions}.
\begin{itemize}
    \item Through detailed circuit-level simulations at scaled process technology nodes, we show that bitline and wordline parasitics are the primary sources of long latency in a crossbar and they create asymmetry in inference latency. With technology scaling, the absolute latency increases and the latency asymmetry becomes increasingly more significant. In addition, different resistance states of a multi-level NVM cell take varying latencies to sense during an inference operation (see Section~\ref{sec:tradeoff}).
    \item We propose to optimize the implementation of synaptic weights on NVM cells such that the resistance state that takes the longest time to sense is programmed on the NVM cell that has the least parasitic delay in a crossbar. This lowers the latency of a crossbar. (see Section~\ref{sec:technology_optimization})
    \item We propose an architectural change of introducing isolation transistors in a crossbar to partition it into regions that can be individually power-gated based on their utilization. In this way, we improve energy efficiency. In addition, by isolating the unutilized region of a crossbar from the active region, parasitics of only the active region contribute to latency, rather than both as in a baseline non-partitioned crossbar architecture. This reduces the latency of a crossbar (see Section~\ref{sec:architecture_optimization}).
    \item We show that our technological and architectural optimizations can only deliver on its latency and energy improvement promises if they are exploited efficiently by the system software. Therefore, we propose a mechanism to expose our proposed design changes to the system-level, allowing the system software to improve both latency and energy when implementing machine-learning inference tasks on hardware (see Section~\ref{sec:software_optimization}).
\end{itemize}

\mr{
We evaluate our design-technology co-optimization approach for a recent neuromorphic hardware using 10 machine learning inference tasks. 
Results show {12\%} reduction in average PE latency and {22\%} lower application energy compared to current state-of-the-art.
}

\mr{
To the best of our knowledge, this is the first work that demonstrates the energy and latency improvement of power gating crossbar-based neuromorphic hardware designs.
}

\section{Design-Technology Tradeoff Analysis}\label{sec:tradeoff}
Without loss of generality, we demonstrate the design-technology tradeoff for an OxRRAM-based neuromorphic PE, where each NVM cell can be programmed to the following four resistance levels (i.e., 2-bit per synapse) -- 1.5 K${\Omega}$, 5.78~K${\Omega}$, 13.6~K${\Omega}$, and 73~K${\Omega}$~\cite{mallik2017design,doevenspeck2021oxrram,chen2020reram,shim2020impact,chiu2010impact}. 
\mr{
Furthermore, we show our analysis for four process technology nodes -- 16nm, 22nm, 32nm, and 45nm, which are obtained from our technology provider. The analysis can be easily extended to other NVM types and also to other process technology nodes.
}

\subsection{Cost-per-Bit Analysis for a Neuromorphic PE}\label{sec:cost_per_bit}
The computer memory industry has thus far been primarily driven by the \textbf{cost-per-bit metric}, which provides the maximum capacity for a given manufacturing cost. As shown in recent works~\cite{mutlu2015research,datacon,lee2013tiered,palp,mutlu2013memory,mneme,hebe}, manufacturing cost can be estimated from the area overhead. To estimate the cost-per-bit of a neuromorphic PE, we investigate the internal architecture of a crossbar and find that a neuron circuit can be designed using 20 transistors and a capacitor~\cite{indiveri2003low}, while an NVM cell is a 1T-1R arrangement with a transistor used as an access device for the cell. Within an $N \times N$ crossbar, there are \ineq{N} pre-synaptic neurons, \ineq{N} post-synaptic neurons, and \ineq{N^2} synaptic cells. The total area of all the neurons and synapses of a crossbar is

\begin{figure}[h!]
	\centering
	\vspace{-5pt}
	\centerline{\includegraphics[width=0.99\columnwidth]{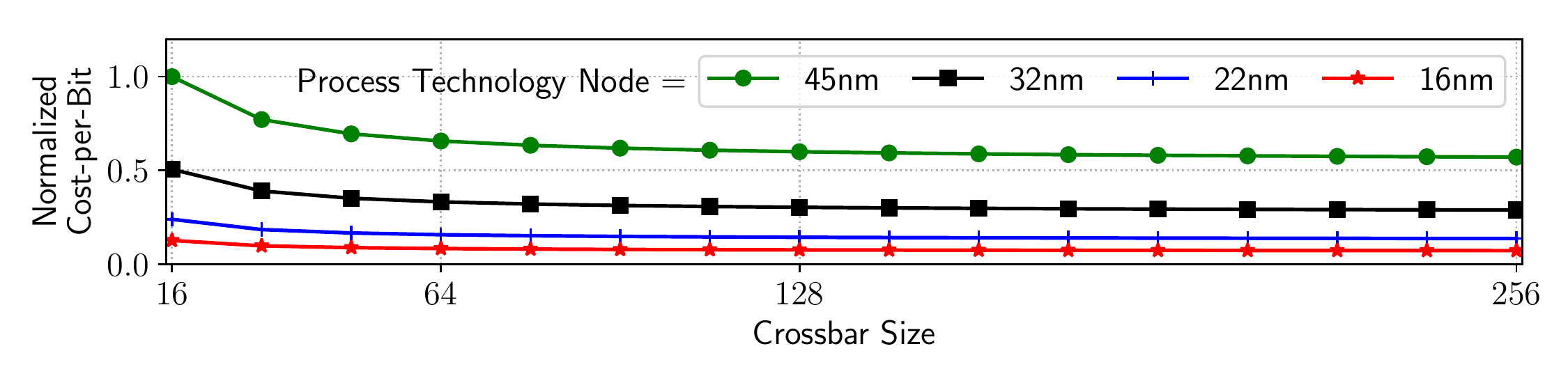}}
	\vspace{-10pt}
	\caption{Cost-per-bit analysis of a crossbar.}
	\vspace{-5pt}
	\label{fig:cost_per_bit}
\end{figure}

\begin{footnotesize}
\begin{eqnarray}
    \label{eq:neuron_synapse_area}
    &\text{neuron area} &= 2N(20T + 1C)\\ \nonumber
    &\text{synapse area} &= N^2(1T + 1R)
\end{eqnarray}
\end{footnotesize}\normalsize
where \ineq{T} stands for transistor, \ineq{C} for capacitor, and \ineq{R} for NVM cell. The total synaptic cell capacity is \ineq{N^2}, with each NVM cell implementing 2-bit per synapse. The total number of bits (i.e., synaptic capacity) in the crossbar is
\begin{equation}
    \label{eq:total_bits}
    \footnotesize \text{total bits} = 2N^2
\end{equation}
Therefore, the cost-per-bit of an \ineq{N}x\ineq{N} crossbar is
\begin{equation}
    \label{eq:cost_per_bit}
    \footnotesize \text{cost-per-bit} = \frac{2N(20T+1C)+N^2(1T+1R)}{2N^2} \approx \frac{F^2(27+2N)}{N},
\end{equation}
where the cost-per-bit is represented in terms of the crossbar dimension \ineq{N} and the feature size \ineq{F}. Equation~\ref{eq:cost_per_bit} provides a \textbf{back-of-the-envelope} calculation of cost-per-bit.
Figure~\ref{fig:cost_per_bit} plots the normalized cost-per-bit for four different process technology nodes, with the crossbar dimension ranging from 16 to 256. We make the following two observations.

 First, the cost-per-bit reduces with increase in the dimension of a crossbar, i.e., larger-sized crossbars can accommodate more bits for a given cost. 
 However, both the absolute latency and latency variation increases significantly for larger-sized crossbars, which increases inference latency and reduces the quality of machine learning inference due to an increase in the ISI distortion (see our analysis in Section~\ref{sec:crossbar_size}).
 Second, the cost-per-bit reduces considerably with technology scaling. This is due to higher integration density at smaller process technology nodes.
 
 \mr{
 The formulation for the cost-per-bit (Equation~\ref{eq:cost_per_bit}) depends on the specific neuron architecture of ~\cite{indiveri2003low} and the one transistor (1T)-based OxRRAM design of~\cite{mallik2017design}. This formulation can be easily extended to other neuron and synapse designs. Furthermore, system designers can use our formulation to configure their neuromorphic hardware, without having to access and plug-in technology-related data.
 }

\subsection{Latency Variation in a Neuromorphic PE}\label{sec:crossbar_size}
\begin{figure}[h!]
	\centering
	\centerline{\includegraphics[width=0.99\columnwidth]{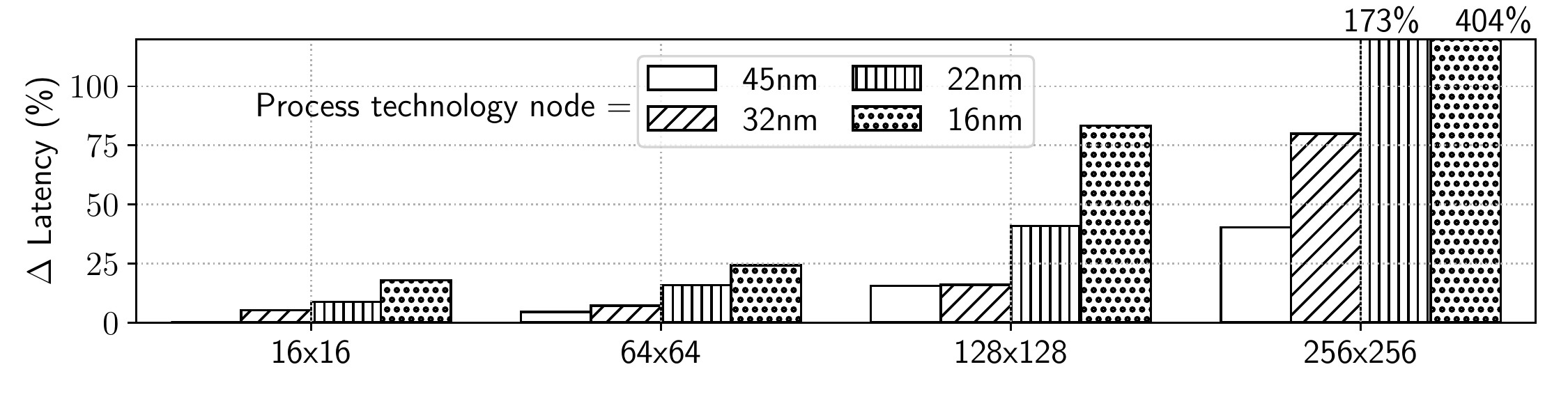}}
	\caption{Variance in latency within a crossbar, \mr{expressed as a fraction of a single spike duration.}}
	\label{fig:crossbar_latency}
\end{figure}

Figure~\ref{fig:crossbar_latency} shows the difference between the best-case and worst-case latency in a crossbar (expressed as a fraction of \ineq{1\mu s} spike duration) for five different crossbar configurations at 45nm, 32nm, 22nm, and 16nm process technology nodes. \mr{See our experimental setup using NeuroXplorer~\cite{neuroxplorer} in Section~\ref{sec:simulation_setup}, which incorporates software, architecture, circuit, and technology.}. All NVM cells are programmed to the HRS state, i.e., 73 K$\Omega$ (see Section \ref{sec:nvm_states} for the dependency on resistance states).

We make two key observations.
First, the latency difference increases with crossbar size due to an increase in the number of parasitic components on current paths.
The average latency difference for $256 \times 256$ crossbar is higher than $16 \times 16$, $64 \times 64$, and $128 \times 128$ crossbar by 16.5x, 13.4x, and 4.5x, respectively. This average is computed across the four process technology nodes. Therefore, smaller-sized crossbars lead to a smaller variation in latency, which is good for performance. However, smaller-sized crossbars also lead to higher cost-per-bit, which we have analyzed in Section~\ref{sec:cost_per_bit}. For most neuromorphic PE designs, $128 \times 128$ crossbars achieve the best tradeoff in terms of latency variation and cost-per-bit~\cite{zhu2018mixed,li2017sneak,he2020towards,paul2021design,dynapse,mallik2017design}.  

Second, the latency difference increases significantly for scaled process technology nodes due to an increase in the value of the parasitic component. 
The average latency difference for 32nm, 22nm, and 16nm process technology nodes is higher than 45nm by 1.3x, 3x, and 6.6x, respectively. 
The unit wordline (bitline) parasitic resistance ranges from approximately \ineq{2.5\Omega} (\ineq{1\Omega}) at {45nm} node to  \ineq{10\Omega} (\ineq{3.8\Omega}) at {16nm} node. The value of these unit parasitic resistances is expected to scale further reaching \ineq{\approx 25\Omega} at {5nm} node~\cite{fouda2017modeling,rakka2020design,fouda2020ir,fouda2019effect,tuli2020rram}. The unit wordline and bitline capacitance values also scale proportionately with technology. Latency variation increases ISI distortion, which degrades the quality of machine learning inference.

\subsection{Varying Latency to Sense NVM Resistance States}\label{sec:nvm_states}
\mr{
The latency (i.e., the delay) on a current path from a pre-synaptic neuron to a post-synaptic neuron within a crossbar is proportion to \ineq{R_{eff}\cdot C_{eff}}, where \ineq{R_{eff}} (\ineq{C_{eff}}) is the effective resistance (capacitance) on the path. This delay increases the time it takes for the membrane potential of a post-synaptic neuron to rise above the threshold voltage causing a delay in spike generation.
}

\begin{figure}[h!]
	\centering
	\vspace{-10pt}
	\centerline{\includegraphics[width=0.99\columnwidth]{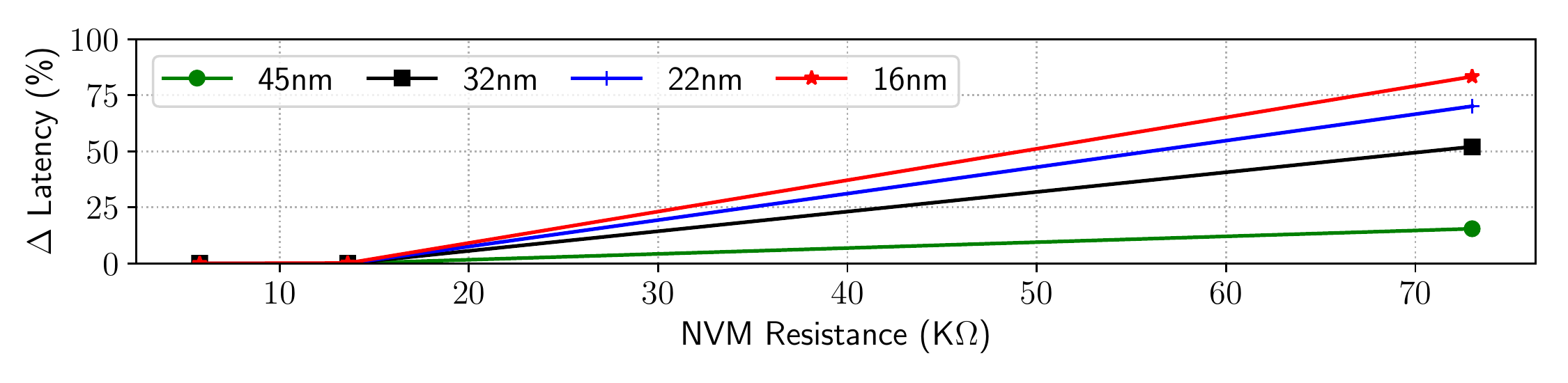}}
	\vspace{-10pt}
	\caption{Latency to sense various NVM resistance states, \mr{expressed as a fraction of a single spike duration.}}
	\vspace{-10pt}
	\label{fig:nvm_state}
\end{figure}

\mr{
The effective resistance on a current path depends on the value of parasitic resistances and the resistance of the NVM cell. We analyze the latency impact due to different resistance states.
}
Figure~\ref{fig:nvm_state} plots the increase in latency (expressed as a fraction of \ineq{1\mu s} spike duration) to sense three NVM resistance states (LRS2, LRS3, and HRS) with respect to LRS1 at 45nm, 32nm, 22nm, and 16nm process technology nodes. These results are for a neuromorphic PE with a $128 \times 128$ crossbar. 

We observe that the latency to sense the HRS state is considerably higher than all three LRS states at all process technology nodes (consistent with \cite{chen2015compact,mallik2017design,yu2014design}). The latency difference increases with technology scaling due to an increase in the size of parasitic components on bitlines and wordlines of a crossbar, which we have analyzed in Section~\ref{sec:crossbar_size}.


\section{Proposed Technological Improvements}\label{sec:technology_optimization}
\begin{figure}[h!]
	\centering
	\centerline{\includegraphics[width=0.99\columnwidth]{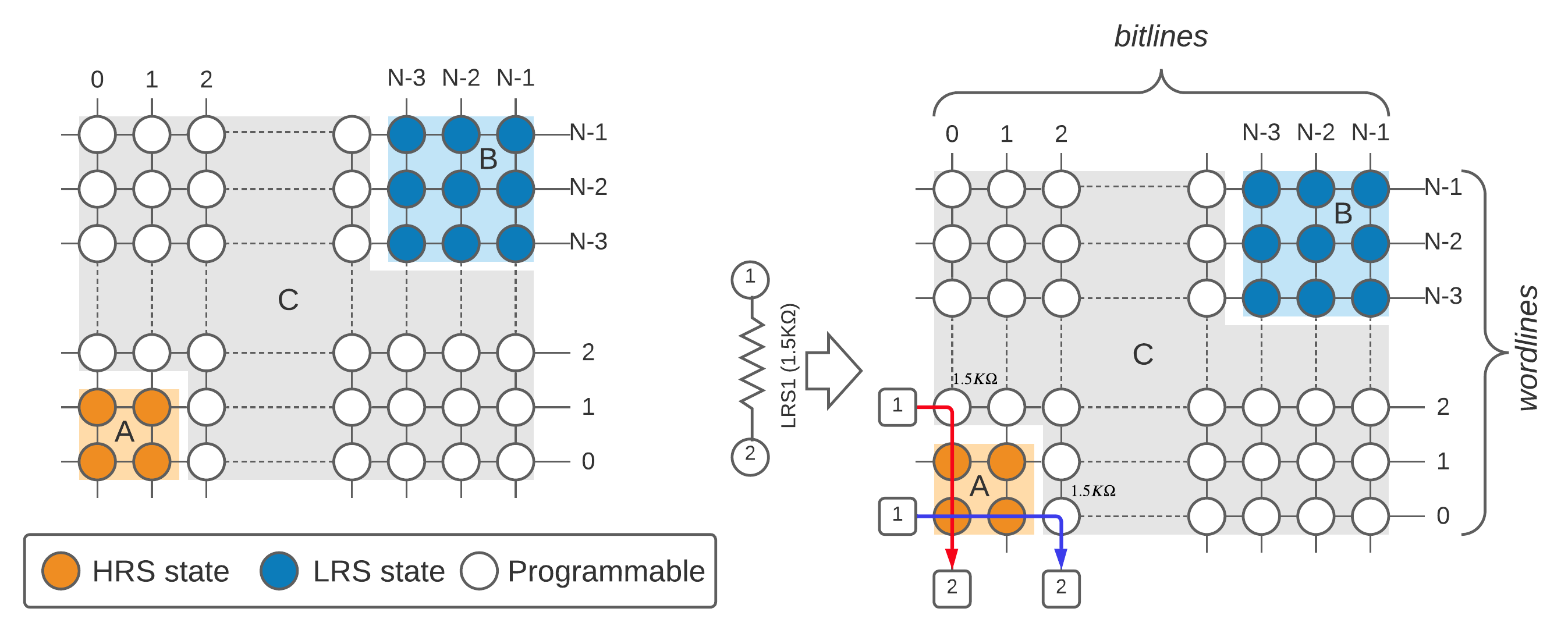}}
	\caption{Our proposed technological change.}
	\label{fig:technological_change}
\end{figure}

Based on the design-technology tradeoff analysis of Section~\ref{sec:tradeoff}, we now present our technology-related optimization.
\mr{
Without loss of generality, we present our optimization for a \ineq{128\times 128} crossbar-based neuromorphic hardware designed at 16nm node.
We exploit the following two observations from Section~\ref{sec:tradeoff}:
}
1) HRS resistance state in an NVM cell takes higher latency to sense than LRS states, and 2) spike propagation latency in a crossbar depends on the number of parasitic components on its current path. 
The left sub-figure of Figure~\ref{fig:technological_change} shows the proposed technological changes. A crossbar is partitioned into three regions. The number of parasitic components on current paths in region A is considerably lower than in the rest of the crossbar. 
\mr{
Therefore, all NVM cells in this region (4 in this example) implement only the HRS resistance state, which takes the longest time to sense.
}
Conversely, NVM cells in region B have longer propagation delay due to higher number of parasitic components. 
\mr{
Therefore, all NVM cells in this region (9 in this example) implement only the LRS resistance state, which takes the shortest time to sense.
}
Finally, all other NVM cells (i.e., those in region C) are programmable, i.e., these cells can implement all four resistance states.
The overall objective is to 
\mr{
balance the latency on different current paths within a crossbar.
}
This minimizes the latency variation in a crossbar, which reduces ISI distortion and improves the quality of machine learning inference tasks.

The right sub-figure shows a pre- and a post-synaptic neuron connected via a synapse that is programmed to the LRS state.
\mr{
The synaptic connection can be implemented on NVM cells in region B (with only LRS1 state) and region C (with programmable states).
}
The figure illustrates two alternative implementations of these neurons.
If the pre-synaptic neuron is implemented on wordline 0, then the post-synaptic neuron \textit{cannot} be implemented on bitlines 0 and 1. This is because NVM cells in region A are all in HRS. In this example, we show the implementation on bitline 2 (see the blue implementation). Conversely, if the post-synaptic neuron is implemented on bitline 0, then the pre-synaptic neuron \textit{cannot} be implemented on wordline 0 and 1 (to avoid using region A). We show the implementation on wordline 2 (see the red implementation).

Formally, the proposed neuromorphic PE is represented by a tuple \underline{\ineq{\langle N,N_h,N_l\rangle}}, where \ineq{N} is the dimension of its crossbar. All NVM cells at crosspoints of wordlines \ineq{0,1,\cdots,N_h-1} and bitlines \ineq{0,1,\cdots,N_h-1} (i.e., region A) can implement only HRS. All NVM cells at crosspoints of wordlines \ineq{N-N_l,N-N_l+1,\cdots,N-1} and bitlines \ineq{N-N_l,N-N_l+1,\cdots,N-1} (i.e., region B) can implement only LRS. All other NVM cells in the PE's crossbar can implement all four resistance states.

\begin{figure}[h!]
	\centering
	\centerline{\includegraphics[width=0.99\columnwidth]{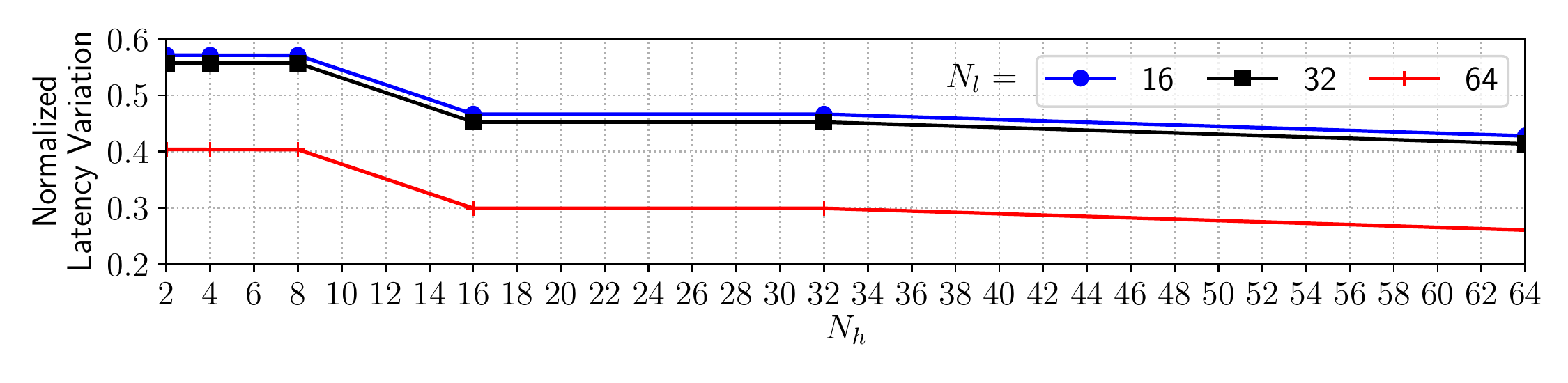}}
	\caption{Latency variation in the proposed crossbar architecture for different settings of $N_h$ and $N_l$.}
	\label{fig:tech_optimize}
\end{figure}

\mr{
Figure~\ref{fig:tech_optimize} plots the variation of latency in the proposed $128 \times 128$ crossbar, normalized to a baseline architecture~\cite{spinemap}, where any NVM cell can be programmed to any of the four resistance states. See Section~\ref{sec:evaluated_approaches} for a description of this baseline architecture and Section~\ref{sec:simulation_setup} for the simulation setup.
The variation in latency is measured as ratio of the best-case and worst-case latency in the crossbar. 
}
The figure reports latency variation for \ineq{N_h} ranging from 2 to 64 with \ineq{N_l} set to 16, 32, and 64. 

\begin{figure}[h!]
	\centering
	\centerline{\includegraphics[width=0.99\columnwidth]{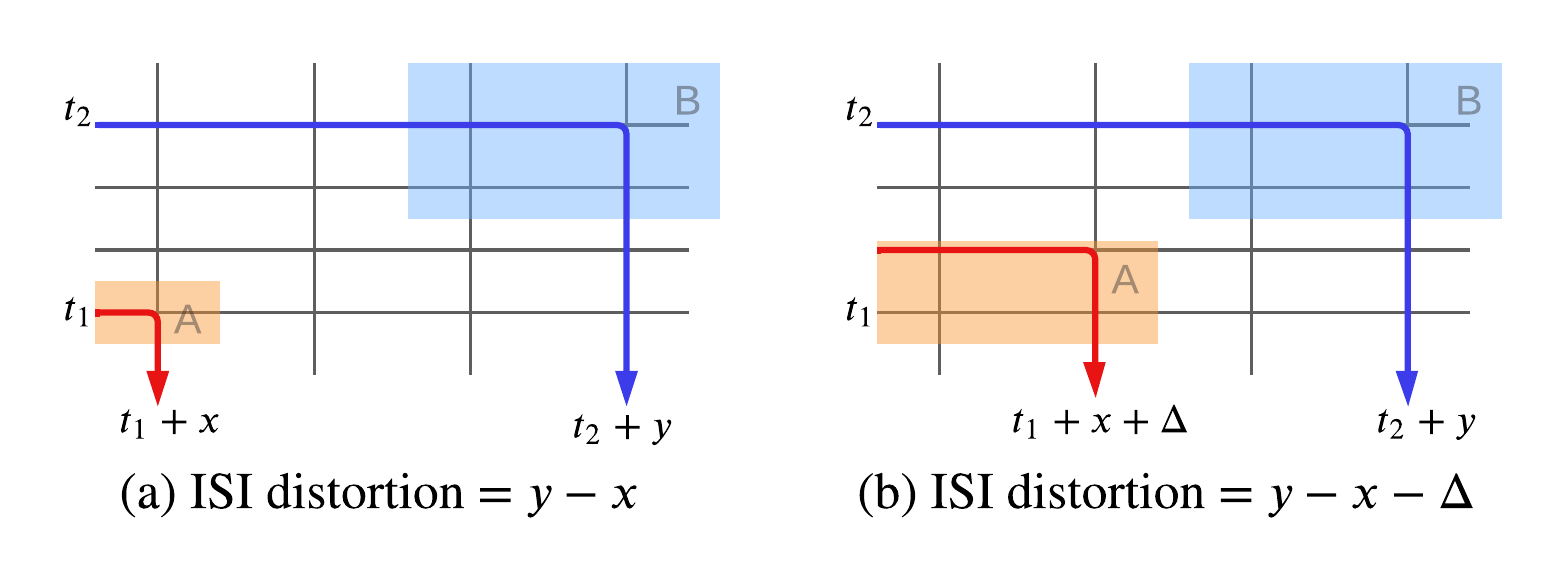}}
	\caption{\minor{ISI improvement due to increase in the size of region A.}}
	\label{fig:isi_var}
\end{figure}

\mr{
We observe that latency variation decreases with an increase in \ineq{N_h}. This is due to an increase in the size of region A, which increases the (worst-case) latency due to an increase in the number of parasitic components on current paths via the HRS state. 
\minor{
However, the (best-case) latency of current paths via the LRS state remains the same. 
Therefore,
the latency variation reduces which improves inference quality by lowering the ISI distortion. 
To illustrate this concept, Figure~\ref{fig:isi_var} provides an example where two synapses are mapped to a \ineq{4\times 4} crossbar.
In Figure~\ref{fig:isi_var}a, the red synapse (in HRS state) is mapped to the bottom left corner of the crossbar, while the blue synapse (in LRS state) to the top right corner. 
The figure shows the timing of two spikes. 
The input spike on the red and blue synapses are \ineq{t_1} and \ineq{t_2}, respectively.
Without loss of generality let \ineq{t_2 > t_1}. 
The ISI of these two spikes is \ineq{t_2 - t_1}.
Due to the delay in current propagation through bitlines and wordlines, these two spikes arrive at the output terminal at different times -- red synapse with a delay of \ineq{x} and blue synapse with a delay of \ineq{y}.
Here, \ineq{y > x}.
Therefore, ISI of the output spikes is \ineq{(t_2 + y) - (t_1 + x)}.
The ISI distortion (difference of ISI between input and output) is 
\begin{equation}
    \label{eq:isi_distortion_ex_a}
    \footnotesize \text{ISI distortion} = \Big((t_2 + y) - (t_1 + x)\Big) - (t_2 - t_1) = y - x
\end{equation}
}
}

\minor{
Figure~\ref{fig:isi_var}b illustrates a scenario where region A is increased to include more cells that are programmed to the HRS state.
The mapping process will map the red synapse using the farthest cell of region A.
The delay on this synapse is \ineq{x+\Delta}, where \ineq{\Delta} is the additional delay due to routing spikes on the red synapse via a longer route compared to that in Figure~\ref{fig:isi_var}a. 
Therefore, ISI of the output spikes is \ineq{(t_2 + y) - (t_1 + x + \Delta)}. 
The ISI distortion is
\begin{equation}
    \label{eq:isi_distortion_ex_b}
    \footnotesize \text{ISI distortion} = \Big((t_2 + y) - (t_1 + x + \Delta)\Big) - (t_2 - t_1) = y - x - \Delta
\end{equation}
}

\minor{
Comparing Equations \ref{eq:isi_distortion_ex_a} \& \ref{eq:isi_distortion_ex_b}, we observe that the ISI distortion reduces due to an increase in the size of region A.
}
\mr{
ISI distortion also reduces with an increase in \ineq{N_l} due to a reduction in the worst-case latency.
We also note that, large \ineq{N_h} may lead to higher average crossbar latency, which impacts real-time performance.
Finally, we see that going from \ineq{N_l = } 16 to 32, there is no significant reduction in the latency variation. Although the size of region B increases with an increase in \ineq{N_l}, we observe only marginal reduction of the best-case latency.
}
Overall, with \ineq{N_h = N_l = 64}, the latency variation is 74\% lower than baseline. 
\mr{
This is chosen based on the tradeoff between latency variation and average latency for \ineq{128\times128} crossbar at 16nm. The tradeoff point can change for other technology nodes and for other crossbar configurations.
}



\subsection{\mr{Reduction in Latency Variation}}
\mr{
To understand the reduction of latency variation within a crossbar as a result of our technological changes, we provide a simple example. Consider there are only two current paths in a crossbar. The parasitic delay on the shortest and longest current paths are \ineq{D} and \ineq{(D+\Delta)}, respectively. The time to sense LRS and HRS NVM states are \ineq{S} and \ineq{(S+\delta)}, respectively. Without any optimization, the worst-case condition is triggered when the HRS state is programmed on the longest path and the LRS state on the shortest path. The minimum and maximum latencies are \ineq{(D+S)} and \ineq{(D+S+\Delta+\delta)}, respectively. The latency variation is \ineq{(\Delta+\delta)}. Using our technology optimization, HRS state is programmed on the shortest path and LRS state on the longest path. The two latencies are \ineq{(D+S+\Delta)} and \ineq{(D+S+\delta)}. The latency variation reduces to \ineq{(|\Delta-\delta|)}. 
}

\mr{
Within a crossbar, there are many current paths (\ineq{N^2} current paths in a \ineq{N\times N} crossbar). The precise reduction in latency variation depends on the specific current paths activated for a synaptic connection, which is controlled during the mapping of a machine learning application to the crossbars of the hardware. In Figure~\ref{fig:tech_optimize}, we show a 74\% reduction comparing only the shortest and the longest paths in a \ineq{128\times128} crossbar. In Section~\ref{sec:latency_variation}, we evaluate the general case considering the mapping process. We report an average 22\% reduction of latency variation.
}

\mr{
Reducing the latency variation helps reduce the ISI distortion, which improves the inference quality. In Section~\ref{sec:inference_quality}, we report an average 4\% increase of inference quality.
}

\subsection{\mr{Impact on Latency}}\label{sec:latency_impact}
\mr{
While latency variation impacts inference quality, the average crossbar latency impacts the real-time performance. To understand the impact of our technological optimization on the average crossbar latency, we consider the same example of two current paths. Consider there are \ineq{m} synapses with LRS states and \ineq{n} synapses with HRS state.
The average latency in the worst-case condition is \ineq{\frac{m\cdot(D+S) + n\cdot(D+S+\Delta+\delta)}{m+n}}. 
Using the technological improvement, the average latency is \ineq{\frac{m\cdot(D+S+\Delta) + n\cdot(D+S+\delta)}{m+n}}. Therefore, the change in latency is \ineq{\left(\frac{n-m}{n+m}\right)\Delta}. This change in latency depends on 1) current paths activated in a crossbar and 2) the value of \ineq{n} and \ineq{m}, i.e., the number of synaptic connections with HRS and LRS states, respectively. In Section~\ref{sec:real_time_performance}, we show an average 3\% reduction of the average crossbar latency for all the evaluated applications.
}


\section{Architectural Enhancements to Neuromorphic PE}\label{sec:architecture_optimization}
\begin{figure}[h!]
	\centering
	\centerline{\includegraphics[width=0.99\columnwidth]{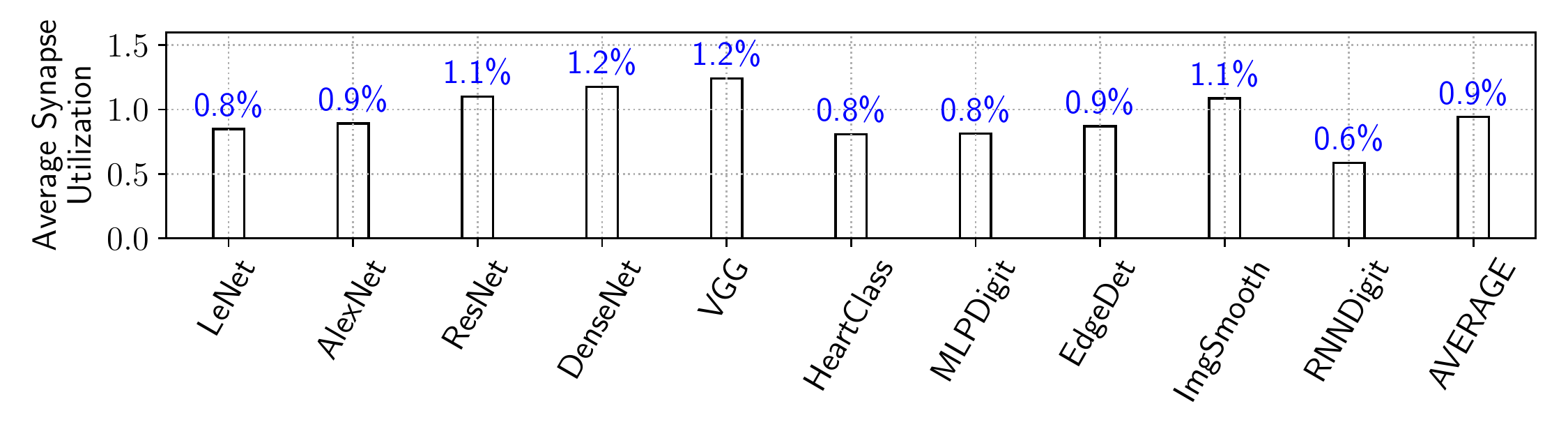}}
	\caption{Average synapse utilization of neuromorphic PEs.}
	\label{fig:utilization}
\end{figure}

To understand the motivation of the proposed architectural changes, Figure~\ref{fig:utilization} reports the average synapse utilization of $128 \times 128$ crossbars in neuromorphic PEs for 10 machine learning models implemented using the spatial decomposition technique of~\cite{esl20}, which is a \textbf{best-effort approach} to improve the utilization of crossbars in a neuromorphic hardware.

\begin{figure}[h!]
	\centering
	\centerline{\includegraphics[width=0.99\columnwidth]{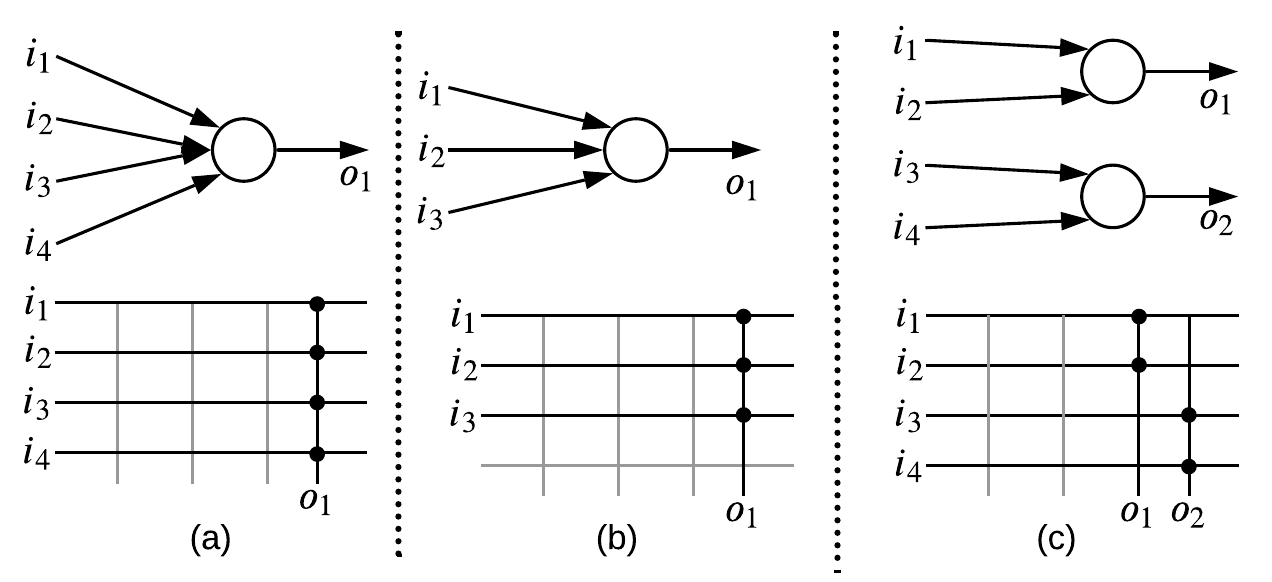}}
	\caption{Implementation of a) one 4-input, b) one 3-input, and c) two 2-input neurons to a $4 \times 4$ crossbar.}
	\label{fig:crossbar_mapping}
\end{figure}

We observe that the average synapse utilization is only 0.9\%. 
\mr{
This is because a crossbar can accommodate only a limited number of pre-synaptic connections per post-synaptic neuron.
}
To illustrate this, Figure~\ref{fig:crossbar_mapping} shows three examples of implementing neurons on a $4 \times 4$ crossbar. The synapse utilization of the three example scenarios are (a) 25\% (4 out of 16), (b) 18.75\% (3 out of 16), and (c) 25\% (4 out of 16). As the crossbar dimension increases, the utilization drops significantly. For instance, if a $128 \times 128$ crossbar is used to implement a single 128-input neuron, i.e., generalization of Fig.~\ref{fig:crossbar_mapping}a, the utilization is only 0.78\% (128 utilized synapses out of a total of $128^2 = 16,384$ synapses). Lower synapse utilization leads to lower energy efficiency.

\begin{figure}[h!]%
    \centering
    \subfloat[Implementing a single neuron function in a partitioned crossbar.\label{fig:single_fn}]{{\includegraphics[width=4.7cm]{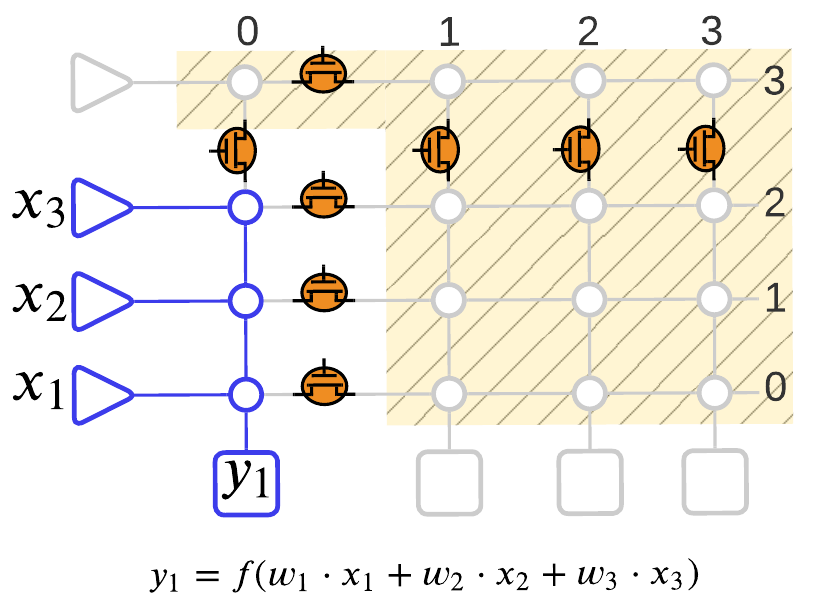} }}%
    \hfill
    \subfloat[Implementing two neuron functions in a partitioned crossbar .\label{fig:two_fns}]{{\includegraphics[width=4.7cm]{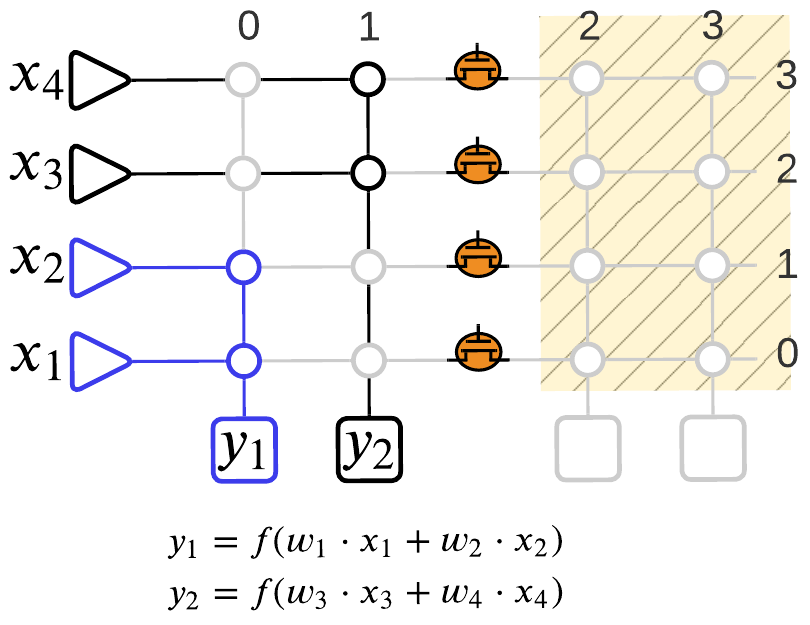} }}%
    \hfill
    \subfloat[Proposed crossbar architecture.\label{fig:proposed_change}]{{\includegraphics[width=4.0cm]{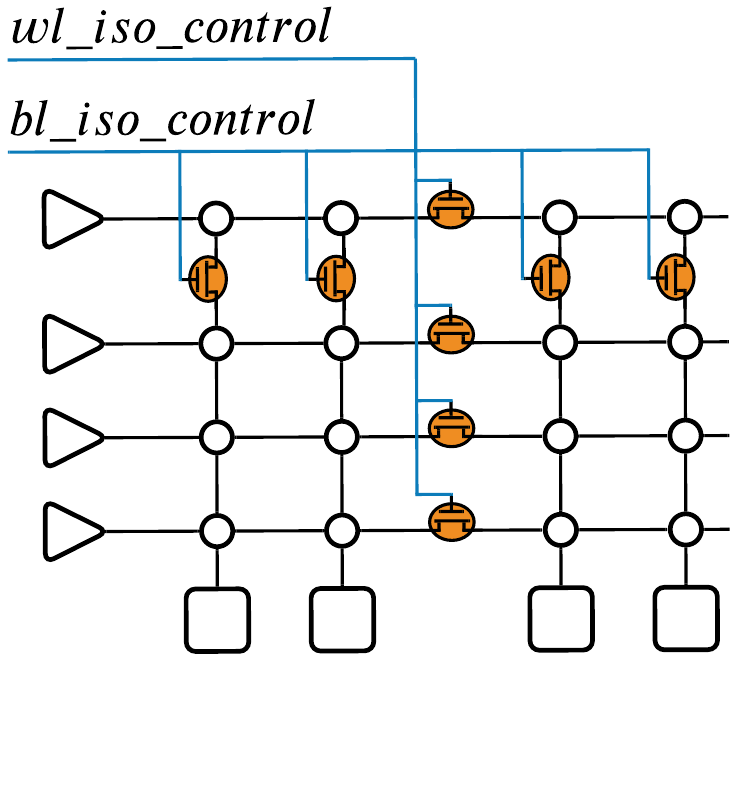} }}%
    \caption{Proposed neuromorphic PE architecture partitioned using isolation transistors.}%
    \vspace{-6pt}
    \label{fig:architectural_change}%
\end{figure}

To improve energy efficiency, we propose to partition a neuromorphic PE into regions that can be dynamically power-gated based on its utilization for a given machine-learning inference task. Figure~\ref{fig:architectural_change} shows the use of isolation transistors in a neuromorphic PE to partition a $4 \times 4$ crossbar into active and unutilized regions. Figure~\ref{fig:single_fn} illustrates the implementation of only a single neuron function \ineq{y_1} in the crossbar. To improve energy efficiency, isolation transistors are needed on every bitline (between wordlines 3 and 4) and on every wordline (between bitlines 1 and 2). Figure~\ref{fig:two_fns} illustrates the implementation of two neuron functions \ineq{y_1} and \ineq{y_2} in the crossbar. In this scenario, isolation transistors are only needed on every wordline (between bitlines 2 and 3). 
To implement inference on a neuromorphic system, each crossbar may have different utilization of its memory cells. Therefore, to improve energy efficiency in every crossbar, isolation transistors are needed on every bitline (and between every pair of wordlines) and on every wordline (and between every pair of bitlines) -- a total of 24 isolation transistors for this example $4 \times 4$ crossbar (in general, \ineqq{2N(N-1)} for an $N \times N$ crossbar). This fine-grained partitioned PE architecture offers flexibility in energy management incorporating crossbar utilization, but leads to a significant increase in the area, latency, and system overhead to 
control isolation transistors.

To overcome these limitations while improving energy efficiency, we enable a coarse-grained partitioning in a crossbar as illustrated in Figure~\ref{fig:proposed_change}. In this example, isolation transistors are inserted selectively on every bitline (between wordlines 3 and 4) and on every wordline (between bitlines 2 and 3). 
This coarse-grained partitioned PE architecture requires a total of 8 isolation transistors (in general, \ineqq{2N} for an $N \times N$ crossbar). To reduce the control overhead, isolation transistors on  wordlines of a crossbar are controlled using a single control signal \texttt{wl\_iso\_ctrl} and those on bitlines using  the signal \texttt{bl\_iso\_ctrl}. Through these two control signals, we enable four distinct configurations of the crossbar, which are summarized in Table~\ref{tab:crossbar_configurations}. 

\begin{table}[h!]
\setlength{\tabcolsep}{2pt}
\caption{Different PE configurations enabled using the two new crossbar control signals.}
\label{tab:crossbar_configurations}
\renewcommand{\arraystretch}{1.0}
\centering
{\fontsize{10}{12}\selectfont
\begin{tabu}{c c | c l l}
    \tabucline[2pt]{-}
    \multicolumn{2}{c|}{\textbf{Crossbar Control}} & \multicolumn{3}{c}{\textbf{Key Parameters}}\\ \hline
    \textbf{wl\_iso\_ctrl} & \textbf{bl\_iso\_ctrl} & \textbf{Dimension} & \textbf{Energy} & \textbf{Latency}\\
    \hline
    \multicolumn{5}{c}{\textcolor{blue}{Baseline PE Architecture}}\\
    \hline
    \multirow{2}{*}{--} & \multirow{2}{*}{--} & \multirow{2}{*}{$4\times4$} & \multirow{2}{*}{$\propto$~4*4} & Best-case: $t_{1,1}$\\
    & & & & Worst-case: $t_{4,4}$\\
    \hline
    \multicolumn{5}{c}{\textcolor{blue}{Proposed Partitioned PE Architecture}}\\
    \hline
    \multirow{2}{*}{0} & \multirow{2}{*}{0} & \multirow{2}{*}{$3\times2$} & \multirow{2}{*}{$\propto$~3*2} & Best-case: $t_{1,1}$\\
    & & & & Worst-case: $t_{3,2}-\Delta$\\
    \hline
    \multirow{2}{*}{0} & \multirow{2}{*}{1} & \multirow{2}{*}{$4\times2$} & \multirow{2}{*}{$\propto$~4*2} & Best-case: $t_{1,1}$\\
    & & & & Worst-case: $t_{4,2}-\Delta + t_{ON}$\\
    \hline
    \multirow{2}{*}{1} & \multirow{2}{*}{0} & \multirow{2}{*}{$3\times4$} & \multirow{2}{*}{$\propto$~3*4} & Best-case: $t_{1,1}$\\
    & & & & Worst-case: $t_{3,4}-\Delta + t_{ON}$\\
    \hline
    \multirow{2}{*}{1} & \multirow{2}{*}{1} & \multirow{2}{*}{$4\times4$} & \multirow{2}{*}{$\propto$~4*4} & Best-case: $t_{1,1}$\\
    & & & & Worst-case: $t_{4,4} + 2\cdot t_{ON}$\\
    \hline
    \tabucline[2pt]{-}
\end{tabu}
}
\end{table}

In a baseline PE architecture, a crossbar dimension is fixed to 4x4. Its static energy is proportional to the number of memory cells, which is 4*4 = 16 in this example. Latency in the crossbar varies from \ineq{t_{1,1}} (nearest cell or best-case) to \ineq{t_{4,4}} (farthest cell or worst-case).

In the proposed partitioned PE architecture, there are four configurations. 

In \underline{configuration `00'}, the crossbar is configured as a 3x2 array with its static energy proportional to 3x2 = 6 memory cells. This is when the unutilized region is power-gated. The best-case latency is \ineq{t_{1,1}} and the worst-case latency is \ineq{t_{3,2}-\Delta}, where \ineq{\Delta} is the reduction in parasitic delay due to shorter bitlines and wordlines.

In \underline{configuration `01'}, the crossbar is configured as a 4x2 array with its static energy proportional to 4x2 = 8 memory cells.
The best-case latency is \ineq{t_{1,1}} and worst-case latency is \ineq{t_{4,2}-\Delta + t_{ON}}, where \ineq{t_{ON}} is the delay of the isolation transistor on current paths.

In \underline{configuration `10'}, the crossbar is configured as a 3x4 array with its static energy proportional to 3x4 = 12 memory cells.
The best-case latency is \ineq{t_{1,1}} and the worst-case latency is \ineq{t_{3,4}-\Delta + t_{ON}}.

In \underline{configuration `11'}, the crossbar is configured as the baseline 4x4 array with its static energy proportional to 4x4 = 16 memory cells.
The best-case latency is \ineq{t_{1,1}} while the worst-case latency is \ineq{t_{4,4} + 2\cdot t_{ON}}. Observe that on the longest current path there are now two isolation transistors, resulting in higher worst-case latency than in the baseline design. 

Our proposed system software (which we discuss in Section~\ref{sec:software_optimization}) minimizes the use of configuration `11', improving both performance and energy efficiency. 

\textbf{Single Control:} The proposed partitioned PE architecture also supports using a single control signal for all the isolation transistors in a crossbar. 
\mr{
When using a single control, only the configurations `00' and `11' are used, implementing a \ineq{3\times 2} and a \ineq{4\times 4} array, respectively.
}

\mr{
To generalize the discussion for an \ineq{N\times N} crossbar,
}
assume that isolation transistors are inserted on every bitline (between wordlines \ineq{P} and \ineq{P+1}) and on every wordline (between bitlines \ineq{Q} and \ineq{Q+1}). Then, the four 
configurations are:
00': a \ineq{P}x\ineq{Q} array; `01': a \ineq{N}x\ineq{Q} array; `10': a \ineq{P}x\ineq{N} array; and `11': a \ineq{N}x\ineq{N} array. 
Formally, \ineq{\langle N,N_h,N_l,P,Q\rangle} represents the proposed partitioned PE architecture.
Equation~\ref{eq:notations} summarizes the notations.

\begin{footnotesize}
\begin{eqnarray}
    \label{eq:notations}
    \langle N\rangle &=& \text{a baseline } N\times N \text{ crossbar}\nonumber\\
    \langle N,N_h,N_l\rangle &=& N\times N \text{ crossbar with tech. enhancement (Sec.~\ref{sec:technology_optimization}})\nonumber\\
    \langle N,N_h,N_l,P,Q\rangle &=& N\times N \text{ crossbar with tech. \& arch. enhancements}\nonumber\\
    && \text{(see Sec.~\ref{sec:technology_optimization} \& \ref{sec:architecture_optimization})}
\end{eqnarray}
\end{footnotesize}

We introduce the following four terminologies: 1) \textbf{expanded mode:} in this mode, a crossbar is operated in configuration `11', 2) \textbf{collapsed mode:} in this mode, a crossbar is operated in configurations `00', `01', and `10', 3) \textbf{collapsed region}, this is the reduced dimension of the crossbar when operating in configurations `00', `01', and `10', and 4) \textbf{far region}, 
\mr{
this is the region of the crossbar excluding the collapsed region.
}

\mr{
In our design methodology, the far region of a crossbar is power-gated using the two control signals at design-time considering the crossbar's utilization. This is achieved during mapping of neurons and synapses to the hardware. Since neuron and synapse mapping does not change during inference, there is no dynamic power management needed. Consequently, there is also no latency and energy overhead involved in switching the far region on/off at run-time.
}

\subsection{\mr{Placing Isolation Transistors in a Crossbar}}\label{sec:iso_placement}
\begin{figure}[h!]
	\centering
	\centerline{\includegraphics[width=0.99\columnwidth]{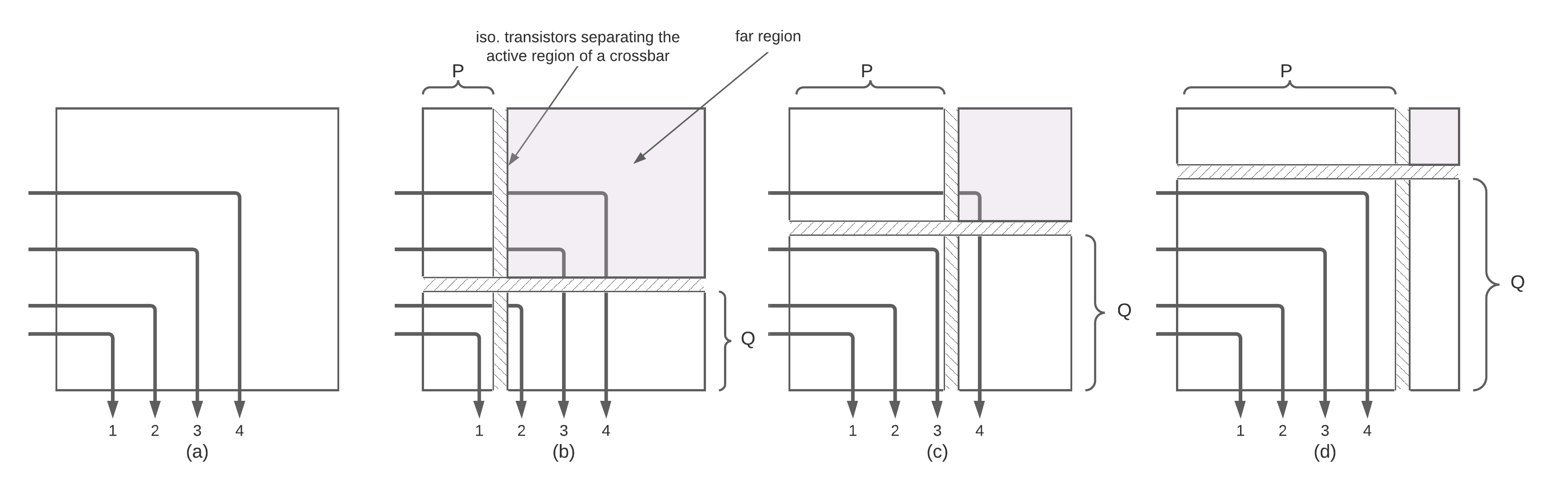}}
	\caption{Placing isolation transistors in a crossbar.}
	\label{fig:iso_placement}
\end{figure}

\mr{
To illustrate the design space exploration involved in placing isolation transistors in a crossbar, Figure~\ref{fig:iso_placement}(a) illustrates a baseline crossbar with four current path that are activated during mapping of neurons and synapses. Figures~\ref{fig:iso_placement}(b)-(d) show three alternative placements of isolation transistors in the crossbar. In Figure~\ref{fig:iso_placement}(b), P and Q values are kept small. The size of the far region is large. In this figure, only two of the current paths (1 \& 2) stays within the collapsed region of the crossbar, while the other two current paths (3 \& 4) traverse via the far region. This means that the latency of paths 3 \& 4 increases due to the delay of the isolation transistors on current paths. Additionally, the far region cannot be power gated, so there is a limited scope for energy reduction using power gating. Increasing P and Q values further (Figure~\ref{fig:iso_placement}(c)), the far region reduces in size as illustrated in the figure. Although three of the four current paths stay in the collapsed region, the far region still cannot be power-gated due to the presence of path 4 in this region. Finally, Figure~\ref{fig:iso_placement}(d) illustrates a possibility where all current paths stay in the collapsed region. The far region can therefore be power-gated. However, because of the small size of the far region, the energy benefits may not be significant. We explore this latency and energy tradeoffs.
}

\begin{figure}[h!]
	\centering
	\centerline{\includegraphics[width=0.99\columnwidth]{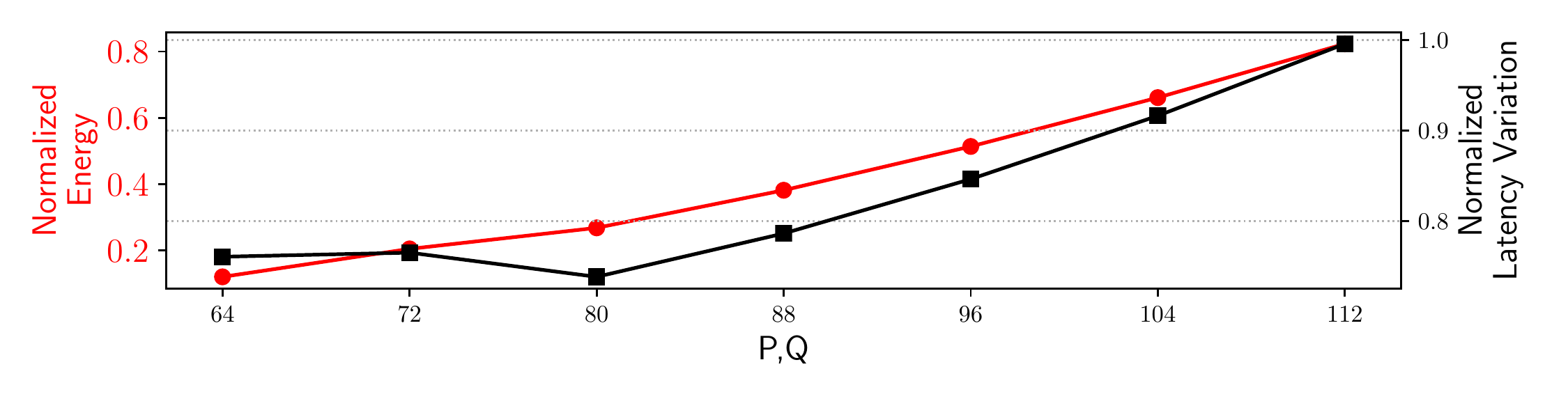}}
	\caption{Selecting P and Q values for the ResNet application.}
	\label{fig:arch_tradeoff}
\end{figure}

Figure~\ref{fig:arch_tradeoff} shows the latency and energy tradeoffs in selecting the values of P and Q for the ResNet inference workload implemented on $128\times 128$ crossbars in a neuromorphic hardware. Latency and energy numbers are normalized to baseline.
We make the following two key observations.

\mr{
First, energy is lower for smaller P and Q values. This is because by reducing P and Q, the size of the collapsed region of a crossbar reduces. 
}
Therefore, there are more memory cells in the far region that can be power-gated to lower energy. 

Second, latency also reduces with a reduction in P and Q values (until P = Q = 80). This is due to shorter bitlines and wordlines of the collapsed region. 
\mr{
However, 
with P = Q = 64 or 72, more clusters of ResNet need crossbars in the expanded mode of operation. 
}
This is because synapses in these clusters can no longer fit onto the reduced dimension of a collapsed crossbar. This increases latency due to isolation transistors on current paths.
For ResNet, P = Q = 80 is the tradeoff point.
\mr{
The tradeoff point is different for different applications. To select a single crossbar configuration that gives good results for all applications, we perform similar analysis for all evaluated applications (see Section~\ref{sec:evaluated_applications}).
Based on such analysis, P = Q = 96 is the selected configuration for the \ineq{128\times128} crossbar at 16nm technology node.
}

\section{Exploiting Technological and Architectural Improvements via the System Software}\label{sec:software_optimization}



\begin{figure}[h!]
	\centering
	\centerline{\includegraphics[width=0.99\columnwidth]{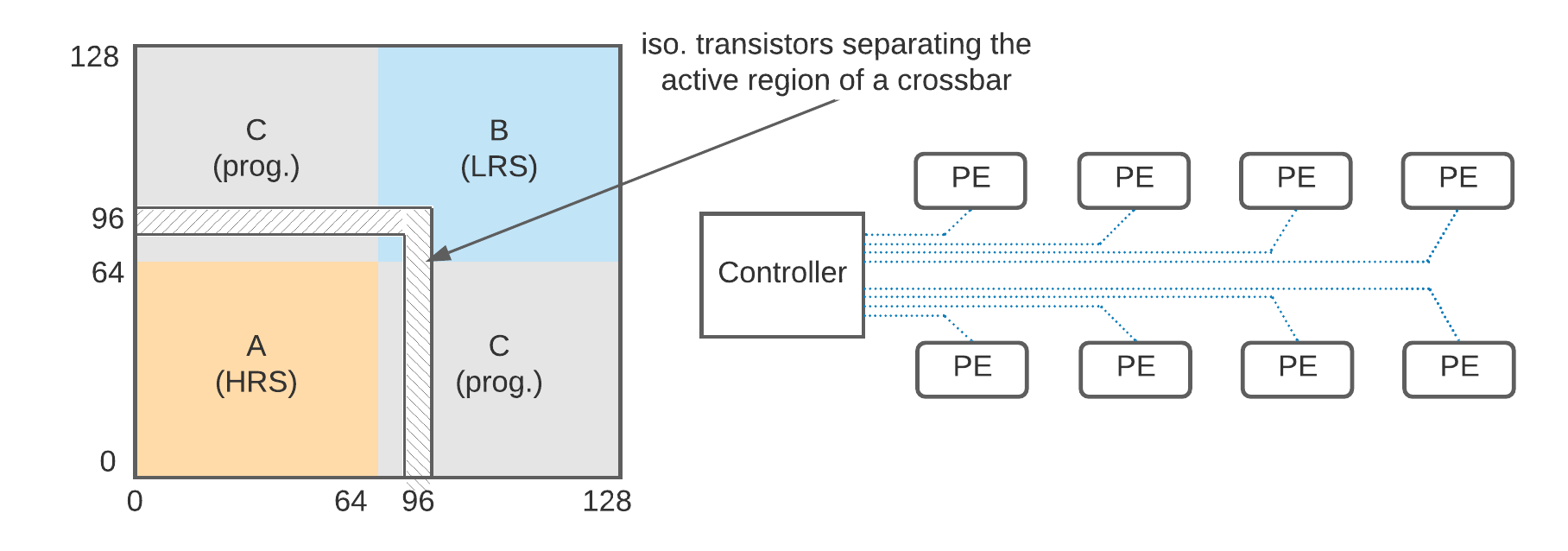}}
	\caption{Final crossbar design using the isolation transistors. The right subfigure shows the control signals generated from the controller when using the proposed partitioned PE architecture in a neuromorphic system.}
	\label{fig:controller}
\end{figure}

To describe the system software,
the left subfigure of Figure~\ref{fig:controller} shows the final crossbar design with isolation transistors that allow each neuromorphic PE to operate in a collapsed or expanded mode. The right subfigure shows control signals for these transistors generated from a centralized controller implemented inside the system software. 

\begin{figure}[h!]
	\centering
	\vspace{-10pt}
	\centerline{\includegraphics[width=0.99\columnwidth]{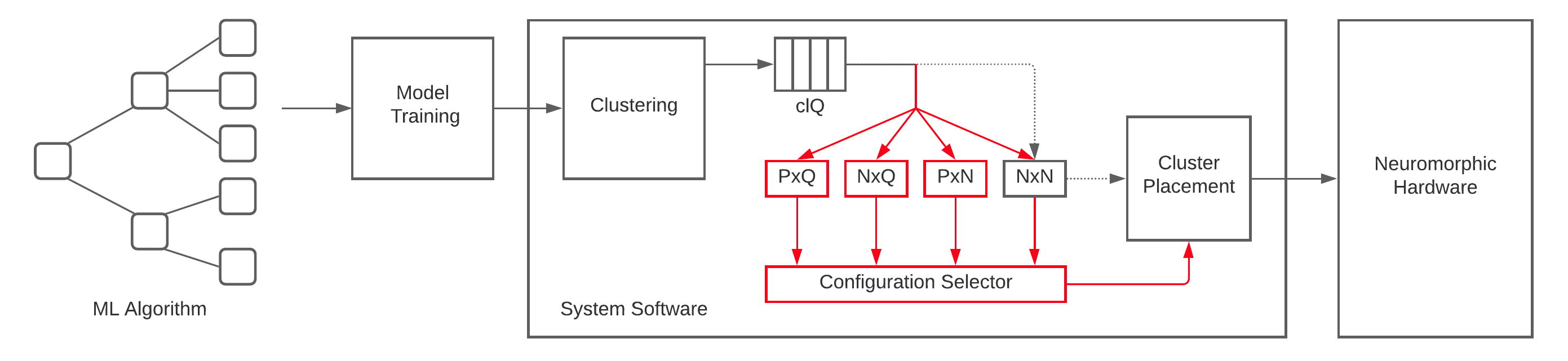}}
	\vspace{-10pt}
	\caption{Proposed system software. All changes are indicated in red.}
	\vspace{-10pt}
	\label{fig:system_software}
\end{figure}

Without loss of generality, Figure~\ref{fig:system_software} shows modifications to the baseline system software~\cite{loihi_mapping} to exploit the proposed design changes. A trained machine learning model is first partitioned to generate clusters, where each cluster can fit onto a crossbar. These clusters are stored in a cluster queue (\texttt{clQ}). 
In the baseline design, each cluster from the clQ is mapped to an $N \times N$ array (exactly replicating the crossbar dimension of the hardware). The mapping is programmed to the hardware using the cluster placement block.
In the proposed design, each cluster of clQ is mapped on four separate arrays -- a $P \times Q$ array, a $N \times Q$ array, a $P \times N$ array, and a $N \times N$ array. 
\mr{
These mappings go to a configuration selection block, which selects the final mapping for the cluster and the configuration of the corresponding PE based on energy-latency tradeoffs. The configuration is programmed to the hardware by configuring the two control signals \texttt{wl\_iso\_ctrl} and \texttt{bl\_iso\_ctrl}. This allows to power-gate the far region of the crossbar. It is important to note that since we power-gate unused resources of a crossbar only at design-time when admitting an application, we minimize the switching overhead. In the future, we will extend this work to also consider dynamic power management by dynamically controlling the isolation transistors. 
}

In selecting the final mapping, the configuration selector first checks to see if a cluster can be mapped to a $P \times Q$ array. If this is possible, then the mapping to the $P \times Q$ array is selected as the final mapping for the cluster, and the corresponding PE is set to operate in configuration `00' (collapsed mode). Otherwise, the configuration selector checks to see if the cluster can be mapped to $N \times Q$ or $P \times N$ array. If so, the corresponding mapping is selected, and the PE is set to operate in configurations `01' or `10', respectively. If the cluster cannot be mapped to either $N\times Q$ or $P\times N$ arrays, the mapping to $N \times N$ array is selected as the final mapping of the cluster with the PE set to operate in configuration `11' (expanded mode).
In this way, the proposed system software uses expanded mode only when it is absolutely necessary to do so. Otherwise, it selects the collapsed region to map synapses, improving both latency and energy.

\section{Evaluation Methodology}\label{sec:evaluation}
\subsection{Simulation Framework}\label{sec:simulation_setup}
We evaluate the proposed design-technology co-optimization approach for OxRRAM-based neuromorphic PEs. 
\mr{
Our simulation framework includes NeuroXplorer~\cite{neuroxplorer}, a cycle-level in-house neuromorphic simulator~\cite{neuroxplorer} with programmable crossbar parameters. We configure this framework to simulate crossbars with parameters listed in Table~\ref{tab:hw_parameters}.
}

\vspace{-10pt}
\begin{table}[h!]
    \caption{Major simulation parameters extracted from~\cite{loihi}.}
	\label{tab:hw_parameters}
	\vspace{-10pt}
	\centering
	{\fontsize{10}{12}\selectfont
		\begin{tabular}{lp{10cm}}
			\hline
			Neuron technology & 16nm CMOS (original design is at 14nm FinFET)\\
			\hline
			Synapse technology & {HfO${}_2$-based OxRRAM}~\cite{mallik2017design}\\
			\hline
			Supply voltage & 1.0V\\
			\hline
			Energy per spike & 23.6pJ at 30Hz spike frequency\\
			\hline
			Energy per routing & 3pJ\\
			\hline
			Switch bandwidth & 3.44 G. Events/s\\
			\hline
	\end{tabular}}
\end{table}
\vspace{-10pt}

Circuit-level simulations are performed with technology parameters from the predictive technology model (PTM)~\cite{zhao2007predictive} and OxRRAM-specific parameters from~\cite{chen2015compact}. 
\mr{
We note that, comparing different chip technologies or recommending one technology node over another is not the focus of this work. Instead, we show that for a given process technology node, design optimizations can reduce energy and latency variations. Furthermore, the proposed design-technology co-optimization methodology can be used by system designers to choose the best technology node for their neuromorphic designs by exploring the energy-performance tradeoffs. 
}

\begin{figure}[h!]
	\centering
	\vspace{-10pt}
	\centerline{\includegraphics[width=0.99\columnwidth]{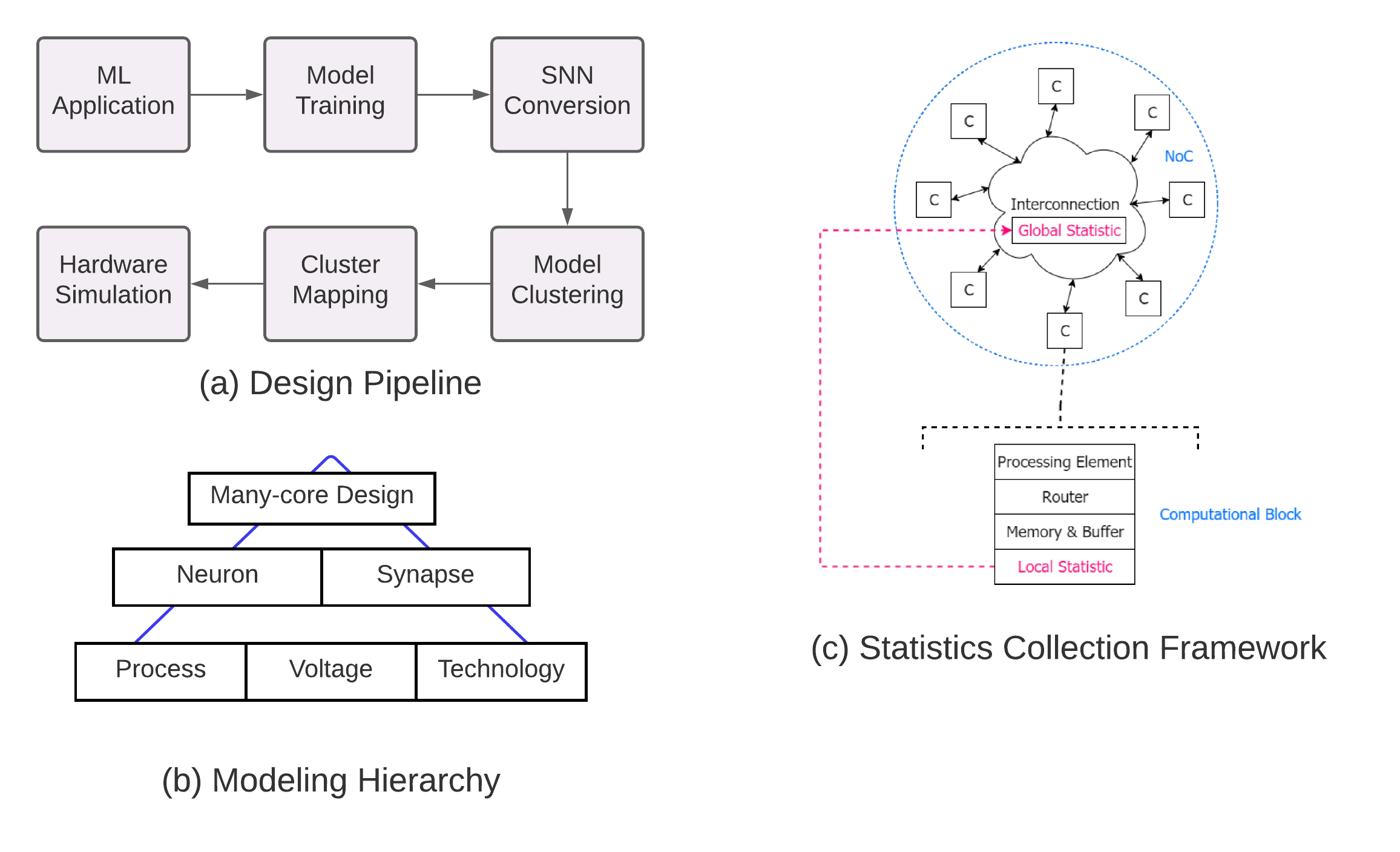}}
	\vspace{-10pt}
	\caption{\mr{Design pipeline using NeuroXplorer.}}
	\vspace{-10pt}
	\label{fig:design_pipeline}
\end{figure}

Neuromorphic simulations are performed on a Lambda workstation, which has AMD Threadripper 3960X with 24 cores, 128 MB cache, 128 GB RAM, and 2 RTX3090 GPUs.
\mr{
Figure~\ref{fig:design_pipeline}(a) shows the design pipeline implemented using NeuroXplorer. A machine learning model is first trained using frameworks such as Keras and PyTorch. Subsequently, the trained model is converted into SNN using~\cite{jolpe18,dfsynthesizer_pp}. The trained model is also simulated using an SNN simulator such as CARLsim~\cite{carlsim}. NeuroXplorer integrates PyCARL~\cite{pycarl}, which allows the SNN model to be simulated using other SNN simulators such as Nengo~\cite{nengo}, Neuron~\cite{neuron}, and Brian~\cite{brian}.
}
Keras~\cite{keras} and CARLsim~\cite{carlsim} both uses the two GPUs to accelerate model training and SNN functional simulation, respectively.

\mr{
The SNN simulated model is clustered using the best-effort technique of~\cite{esl20}, which maximizes cluster utilization. Clusters of the SNN are mapped to the hardware using the SpineMap technique~\cite{spinemap}. Finally, we perform cycle-accurate simulation of the clusters using NeuroXplorer~\cite{neuroxplorer}.
}

\mr{
Figure~\ref{fig:design_pipeline}(b) shows the modeling hierarchy of the simulator. At the highest level is the many-core design, which is a tile-based architecture, similar to Loihi~\cite{loihi}. Each PE consists of a crossbar, which is an organization of neurons and synapses. A neurons is modeled using~\cite{indiveri2003low} and a synaptic circuit using~\cite{mallik2017design}. At the lowest level are the technology models (see Table~\ref{tab:hw_parameters}).
}

\mr{
Finally, Figure~\ref{fig:design_pipeline}(c) shows the statistics collection framework in NeuroXplorer. It facilitates global statistics collection, where spike arrival times are recorded for each PE (shown as C in the figure). These spike times are then used to compute the ISI distortion (see Appendix~\ref{sec:isi_distortion}).
}

\subsection{Power Consideration for Isolation Transistors}
The {additional power required to control the isolation transistors} when accessing the RRAM cells in the far region is approximately 3x that of raising a wordline, since raising a wordline requires driving one access transistor per bitline, while accessing the RRAM cells in the far region requires driving two isolation and one access transistor per bitline. \mr{The power overhead for accessing RRAM cells in the collapsed mode `01' and `10' is approximately 2x (one isolation and one access transistor)~\cite{mneme,hebe,lee2013tiered}.} The energy numbers reported in Section~\ref{sec:energy_efficiency} incorporates these overheads.

\subsection{Evaluated Workloads}\label{sec:evaluated_applications}
We select 10 machine learning inference programs that are representative of three most commonly-used neural network classes: convolutional neural network (CNN), multi-layer perceptron (MLP), and recurrent neural network (RNN).
Table~\ref{tab:apps} summarizes the topology, number of neurons and synapses, number of spikes per image, and baseline quality of these applications on hardware.

\vspace{-10pt}
\begin{table}[h!]
	\renewcommand{\arraystretch}{1.2}
	\setlength{\tabcolsep}{2pt}
	\caption{Applications used to evaluate the proposed approach.}
	\label{tab:apps}
	\vspace{-10pt}
	\centering
	\begin{threeparttable}
	{\fontsize{10}{12}\selectfont
		\begin{tabular}{ccc|ccc|cc}
			\hline
			\multicolumn{3}{c|}{} & \multicolumn{3}{c|}{} & \textbf{Baseline} & \textbf{\mr{Obtained}}\\
			\textbf{Class} & \textbf{Applications} &
			\textbf{Dataset} &
			\textbf{Neurons} & \textbf{Synapses} & \textbf{Avg. Spikes/Frame} & \textbf{Quality} & \textbf{\mr{Quality}}\\
			\hline
			\multirow{6}{*}{CNN} & LeNet & CIFAR-10 & 80,271 & 275,110 & 724,565 & 86.3\% & \mr{87.1\%}\\
			& AlexNet & CIFAR-10 & 127,894 & 3,873,222 & 7,055,109 & 66.4\% & \mr{66.9\%}\\
			& ResNet & CIFAR-10 & 266,799 & 5,391,616 & 7,339,322 & 57.4\% & \mr{58.0\%}\\
			& DenseNet & CIFAR-10 & 365,200 & 11,198,470 & 1,250,976 & 46.3\% & \mr{46.5\%}\\
			& VGG & CIFAR-10 & 448,484 & 22,215,209 & 12,826,673 & 81.4 \% & \mr{81.6\%}\\
			& HeartClass~\cite{das2018heartbeat} & Physionet & 170,292 &  1,049,249 & 2,771,634 & 63.7\% & \mr{63.9\%}\\
			\hline
			\multirow{3}{*}{MLP} & MLPDigit & MNIST & 894 & 79,400 & 26,563 & 91.6\% & \mr{96.4\%}\\
			& EdgeDet \cite{carlsim} & CARLsim & 7,268 &  114,057 & 248,603 & SSIM = 0.89 & \mr{0.99}\\
			& ImgSmooth \cite{carlsim} & CARLsim & 5,120 & 9,025 & 174,872 & PSNR = 19 & \mr{22.2}\\
			\hline
 			RNN & RNNDigit \cite{Diehl2015} & MNIST & 1,191 & 11,442 & 30,508 & 83.6\% & \mr{83.7\%}\\
			\hline
	\end{tabular}}
	\end{threeparttable}
\end{table}
\vspace{-10pt}

\subsection{Evaluated Approaches}\label{sec:evaluated_approaches}
We evaluate the following techniques.
\begin{itemize}
    \item \textit{\underline{Baseline}}~\cite{spinemap}. The Baseline approach first clusters a machine-learning inference model to minimize the inter-cluster spike communication. Clusters are then mapped to neuromorphic PEs of the hardware with synapses of each cluster implemented on memory cells of a crossbar without incorporating latency variation. Neuromorphic PEs are not optimized to reduce latency variation, \mr{i.e., any resistance state (LRS or HRS) can be programmed on any current path (long or short).}
    \mr{Unused crossbars are power-gated to reduce energy consumption. This is the coarse-grained power management technique implemented in many state-of-the-art many-core neuromorphic designs such as Loihi~\cite{loihi}, DYNAPs~\cite{dynapse}, and \mubrain{}~\cite{sentryos}.}
    
    \item \textit{\underline{Baseline + Design Changes}}. This is the Baseline mapping approach implemented on the proposed latency-optimized partitioned neuromorphic PE design. \mr{In the proposed design, HRS state, which takes long time to sense, is used only on shorter current paths, ones that have lower parasitic delays. Similarly, LRS state is used only on loner current paths. In addition to coarse-grained power management, we facilitate power gating at a finer granularity in the proposed design. Specifically, by controlling the isolation transistors, we power-gate unused resources within each crossbar.}
    
    \item \textit{\underline{Proposed}}. This is the proposed solution where the system software is optimized to exploit the design changes.
\end{itemize}

\section{Results and Discussions}\label{sec:results}

\subsection{Energy Efficiency}\label{sec:energy_efficiency}
\begin{figure}[h!]
	\centering
	\centerline{\includegraphics[width=0.99\columnwidth]{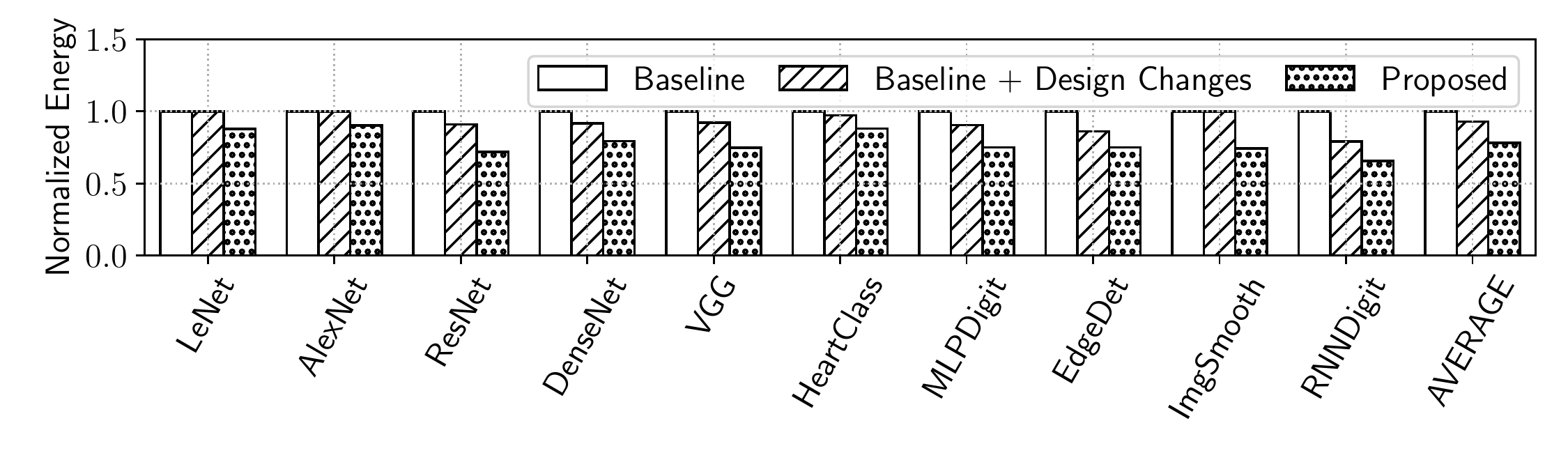}}
	\caption{Energy consumption normalized to Baseline.}
	\label{fig:energy_efficiency}
\end{figure}

Figure~\ref{fig:energy_efficiency} plots the energy efficiency of the evaluated techniques normalized to Baseline. We make the following two key observations.

First, with the proposed design changes, energy reduces by only 7\% compared to Baseline.
This is because, both in Baseline and Baseline with the proposed design changes, synapses of a cluster are implemented randomly on NVM cells of a crossbar causing them to be distributed across the crossbar dimension. Therefore, there remains a limited scope to collapse the crossbar and use power-gating to save energy.
Second, the proposed design-technology co-optimization approach has the lowest energy (22\% lower than Baseline and 16\% lower than Baseline with the proposed design changes). This improvement is due to the proposed system software, which exploits the design changes in implementing machine learning inference on neuromorphic PEs. In particular, synapses are implemented to maximize the utilization of the collapsed region in each crossbar of the hardware. If all of a cluster's synapses fit into the collapsed region, then the far region can be isolated from the collapsed region using isolation transistors and power-gated to save energy.

\subsection{Latency Variation}\label{sec:latency_variation}
\begin{figure}[h!]
	\centering
	\centerline{\includegraphics[width=0.99\columnwidth]{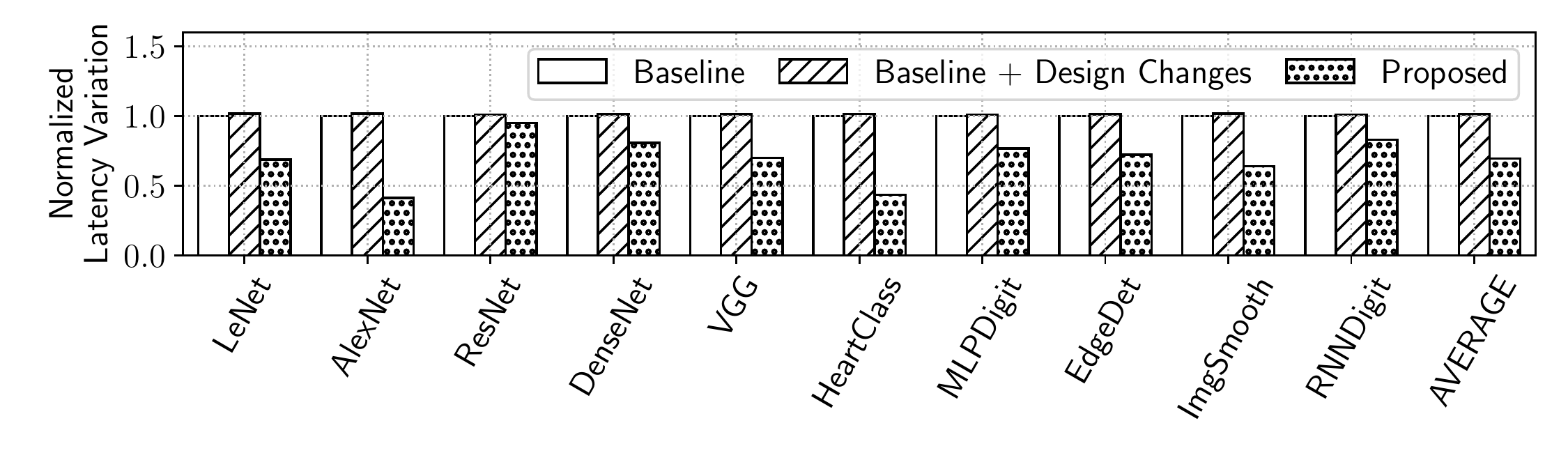}}
	\caption{Latency variation normalized to Baseline.}
	\label{fig:latency_variation}
\end{figure}

Figure~\ref{fig:latency_variation} plots the latency variation normalized to Baseline. We make the following three key observations. 

First, with the proposed design changes, latency variation increases compared to Baseline by average 1\%. This is because of the increase in latency associated with the delay of isolation transistors on current paths. Second, the latency variation using the proposed approach is 30\% lower than Baseline and 32\% lower than Baseline with the proposed design changes. The reason for these improvements is three fold -- 1) optimizing NVM resistance states in a crossbar such that the state that takes the longest time to sense is programmed on current paths that have the least propagation delay, 2) isolating the collapsed region of a crossbar from the far region, to reduce current propagation delay, and 3) exploiting these changes during the implementation of a machine learning inference using the proposed system software, which uses the far region of a crossbar only when it is absolutely necessary to do so. Otherwise, it improves both latency and energy by operating the crossbar in the collapsed mode.

\mr{
Finally, the latency variation using the proposed approach varies across different applications. This is because the proposed approach exploits the latency and energy tradeoffs differently for different applications. The latency variation is similar to the Baseline for ResNet, while it is significantly lower than the Baseline for HeartClass.} 

\mr{
Using the results from Sections~\ref{sec:energy_efficiency} and \ref{sec:latency_variation}, we conclude that the proposed approach introduces maximum gain for applications where the latency and energy tradeoffs can be better exploited. For all other applications, it either minimizes energy or minimizes latency variation.
}

\subsection{\mr{Real-time Performance}}\label{sec:real_time_performance}
\begin{figure}[h!]
	\centering
	\centerline{\includegraphics[width=0.99\columnwidth]{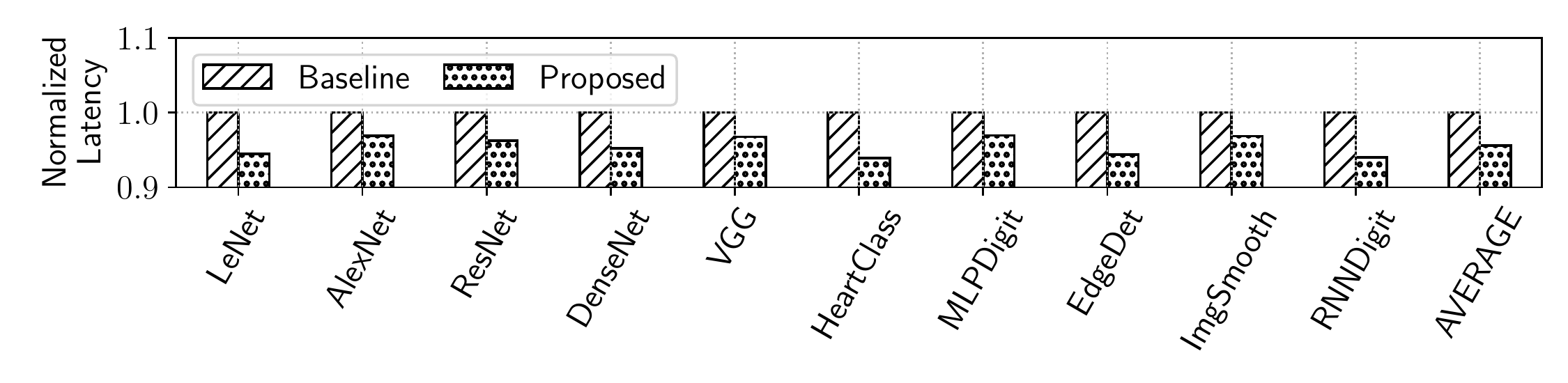}}
	\caption{\mr{Crossbar latency normalized to Baseline.}}
	\label{fig:latency}
\end{figure}

\mr{
One of the key hardware performance metrics for neuromorphic computing is real-time performance, which is a function of the crossbar latency. To evaluate real-time performance, Figure~\ref{fig:latency} plots the crossbar latency of the proposed approach and the Baseline for the evaluated applications. Results are normalized to the Baseline.
}

\mr{
We observe that the crossbar latency using the proposed approach is on average 4.5\% lower than the Baseline. This reduction is because the proposed approach places synapses with the HRS state on shorter current paths, which lowers the overall spike latency on those synapses. We have elaborated this in Section~\ref{sec:latency_impact}.
}


\subsection{Inference Quality}\label{sec:inference_quality}
\begin{figure}[h!]
	\centering
	\centerline{\includegraphics[width=0.99\columnwidth]{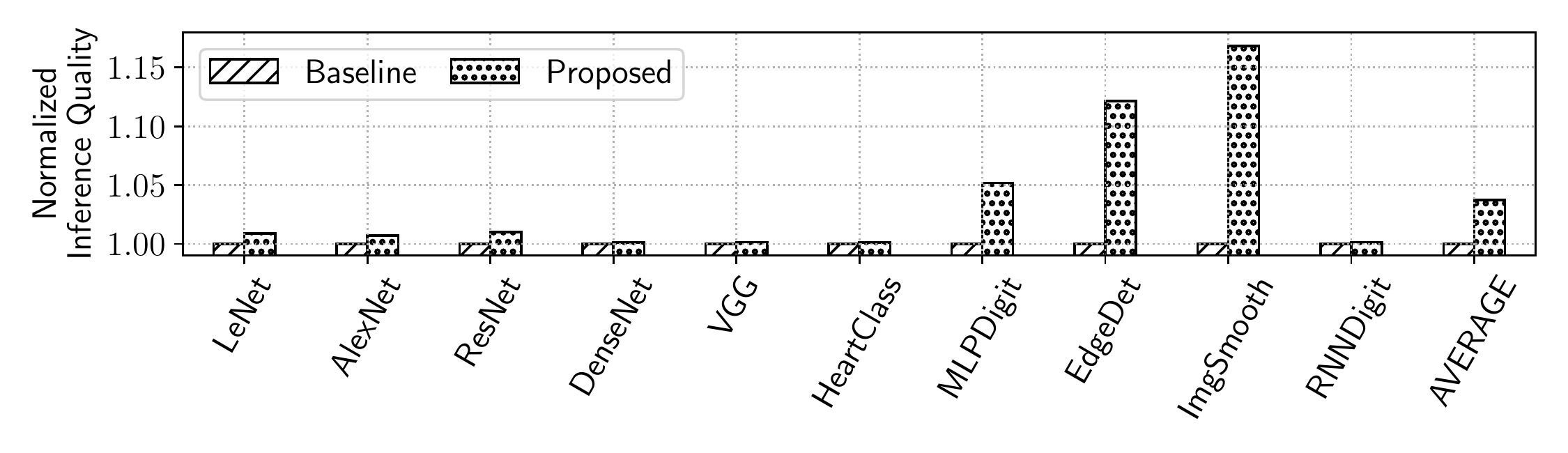}}
	\caption{Inference quality normalized to Baseline.}
	\label{fig:quality}
\end{figure}

Figure~\ref{fig:quality} shows the improvement in inference quality using the proposed approach, normalized to Baseline. We observe that the image quality improves by an average of 4\%. This is due to the reduction in ISI distortion caused by a reduction of the latency variation in neuromorphic PEs using the proposed changes, which we have analyzed in Section~\ref{sec:latency_variation}. In addition, the improvement of inference quality with PSNR and SSIM metrics for EdgeDet and ImgSmooth is higher than other inference tasks with accuracy metrics. This is because PSNR and SSIM metrics are computed on individual images where we see a large improvement in quality. For accuracy-based tasks, 
we observe that feature representation in hidden layers of these models changes due to ISI distortion, but not all such changes lead to misclassification. So the accuracy of these inference tasks is comparable to Baseline.

\subsection{Single vs. Double Control Design}
\begin{figure}[h!]
	\centering
	\centerline{\includegraphics[width=0.99\columnwidth]{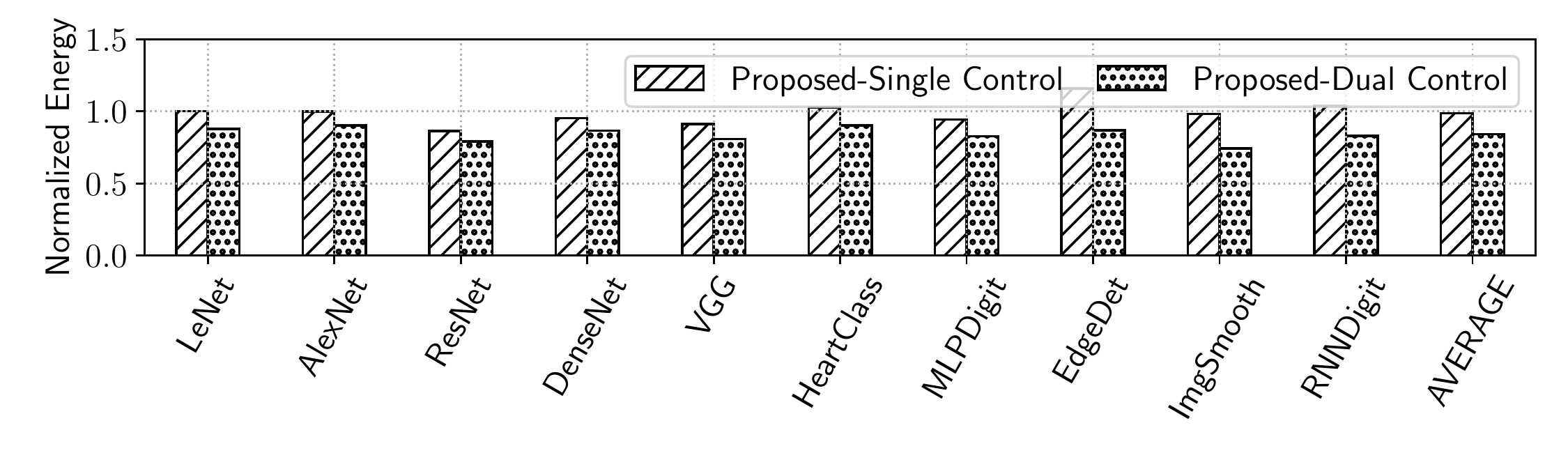}}
	\caption{Partitioned PE architecture with single and double control.}
	\label{fig:config}
\end{figure}

Figure~\ref{fig:config} plots the energy efficiency of the proposed design with single control signal and the default, which uses two control signals for each PE. We observe that using single control, energy reduces by only 2\% compared to Baseline. This is because most crossbars are operated in the expanded mode due to limited scope to collapse the crossbar. Our default design leads to 14.4\% lower energy than with single control. This is because in the default design, a crossbar can be collapsed along X- and Y- dimensions independently, leading to three collapsed array configurations. Therefore, the system software has a higher probability to use the collapsed mode, leading to a reduction in energy.

\subsection{Die Area Analysis}
Adding an isolation transistor to the bitline increases the height of the crossbar, whereas that on the wordline increases the width. Without the isolation transistors, the height of a baseline crossbar is equal to the sum of height of the memory cells and the sense-amplifier, while the width is equal to the sum of the width of the memory cells. \mr{For RRAM-based neuromorphic PEs, a sense amplifier in the peripheral circuit and a isolation transistor is approximately 384x and 9.6x taller than an individual RRAM cell, respectively~\cite{mallik2017design,xu2011design,chen2016design}.} In terms of width, an isolation transistor is only 1.3x wider than an RRAM cell. Therefore, for a crossbar with 128 RRAM cells per bitline and wordline (i.e., $128 \times 128$ array), the overhead along the height of the crossbar is \ineq{\frac{9.6}{384 + 128} = 1.83\%}, and the overhead along the width of the crossbar is \ineq{\frac{1.3}{128} = 1.01\%}.


\section{Conclusions}\label{sec:conclusions}
We present a design-technology co-optimization approach to implement energy-efficient machine-learning inference on NVM-based neuromorphic processing elements (PEs). First, we optimize the NVM resistance state such that the state that takes the longest time to sense is placed on current paths with fewer parasitics, and hence incurs lower propagation delay, and vice versa. Second, we use isolation transistors to partition a PE into collapsed and far regions such that the NVM cells of the far region can be opportunistically power-gated to save both energy and latency. Finally, we use the system software to exploit the design changes, maximizing the utilization of the collapsed region of each PE in the hardware. Our system software uses the far region only when it is absolutely necessary to do so, otherwise it improves both latency and energy by operating the PE in the collapsed mode. We evaluate our design-technology co-optimization approach for a state-of-the-art neuromorphic architecture. Evaluations with different machine-learning inference tasks show that the proposed approach improves both latency and energy without incurring significant cost-per-bit.

\begin{acks}
This work is supported by 1) U.S. Department of Energy under Award Number DE-SC0022014, 2) the National Science Foundation Award CCF-1937419 (RTML: Small: Design of System Software to Facilitate Real-Time Neuromorphic Computing) and 3) the National Science Foundation Faculty Early Career Development Award CCF-1942697 (CAREER: Facilitating Dependable Neuromorphic Computing: Vision, Architecture, and Impact on Programmability).
\end{acks}

\bibliographystyle{IEEEtranSN}
\bibliography{commands,disco,external}

\appendix
\section{Spiking Neural Networks}
\begin{figure}[h!]
	\centering
	\centerline{\includegraphics[width=0.99\columnwidth]{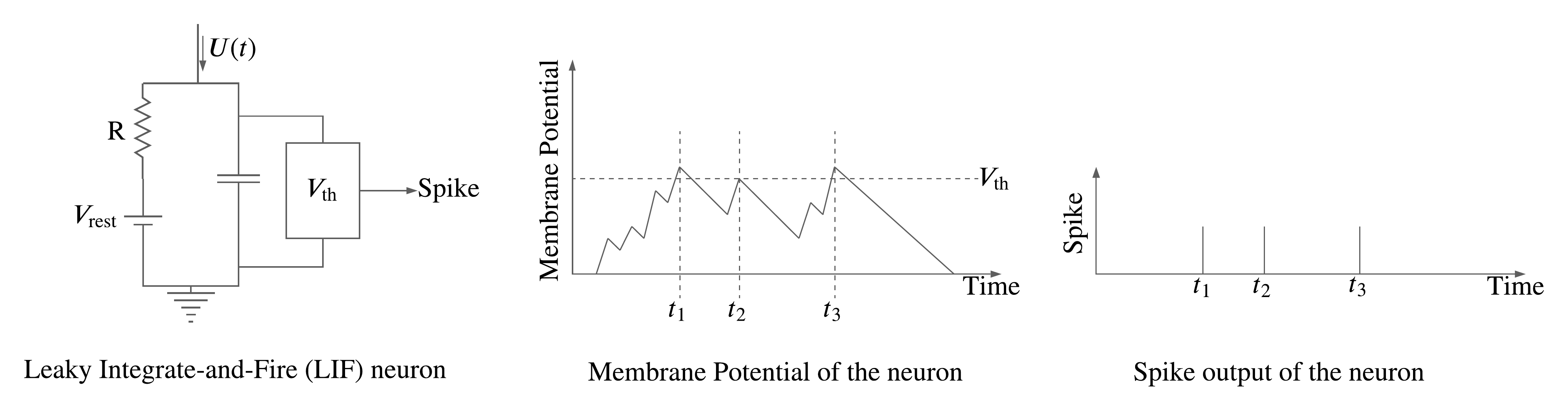}}
	\caption{A leaky integrate-and-fire (LIF) neuron with current input $U(t)$ (left). The membrane potential over time of the neuron (middle). The spike output of the neuron representing its firing time (right).}
	\label{fig:lif}
\end{figure}

Spiking Neural Networks (SNNs) enable powerful computations due to their spatio-temporal information encoding capabilities~\cite{maass1997networks}. 
\mr{
An SNN consists of neurons, which are connected via synapses. A neuron can be implemented as an integrate-and-fire (IF) logic, which is illustrated in Figure~\ref{fig:lif} (left). Here, an input current \ineq{U(t)} (i.e., spike from a pre-synaptic neuron) raises the membrane voltage of the neuron. When this voltage crosses a threshold \ineq{V_{th}}, the IF logic emits an output spike, which propagates to is post-synaptic neuron. 
Figure~\ref{fig:lif} (middle) illustrates the membrane voltage of the IF neuron due to an input spike train. The moment of threshold crossing is illustrated in Figure~\ref{fig:lif} (right). These are the firing times of the output spike train of the neuron.
}

\mr{
SNNs can implement many machine learning approaches such as supervised learning, unsupervised learning, reinforcement learning, and lifelong learning.
We focus on
}
supervised machine learning,
where an SNN is pre-trained with representative data. Machine learning \textbf{inference} refers to feeding live data points to this trained SNN to generate the corresponding output.

\section{Quality of Inference}\label{sec:isi_distortion}
The \textbf{quality} of machine learning inference can be expressed in terms of accuracy~\cite{jolpe18}, Mean Square Error (MSE)~\cite{HeartEstmNN}, Peak Signal-to-Noise Ratio (PSNR)~\cite{carlsim}, and Structural Similarity Index Measure (SSIM)~\cite{hore2010image}.
\mr{
While accuracy is commonly used for assessing the quality of supervised learning, e.g., using Convolution Neural Networks (CNNS), there are also applications such as edge detection, where the quality is assessed using other metrics such as SSIM.
In our prior work~\cite{spinemap}, we have shown that these quality metrics are a function of the inter-spike interval (ISI) between neurons. Therefore, any deviation of ISI (called ISI distortion) from its trained value may lead to quality loss. To describe ISI,
}
let  \ineq{\{t_1,t_2,\cdots,t_{K}\}} denote a neuron's firing times in the time interval \ineq{[0,T]}, the average ISI of this spike train is

\begin{equation}
    \label{eq:isi}
    \footnotesize \mathcal{I} = \sum_{i=2}^K (t_i - t_{i-1})/(K-1).
\end{equation}

\begin{figure}[h!]
	\centering
	\centerline{\includegraphics[width=0.99\columnwidth]{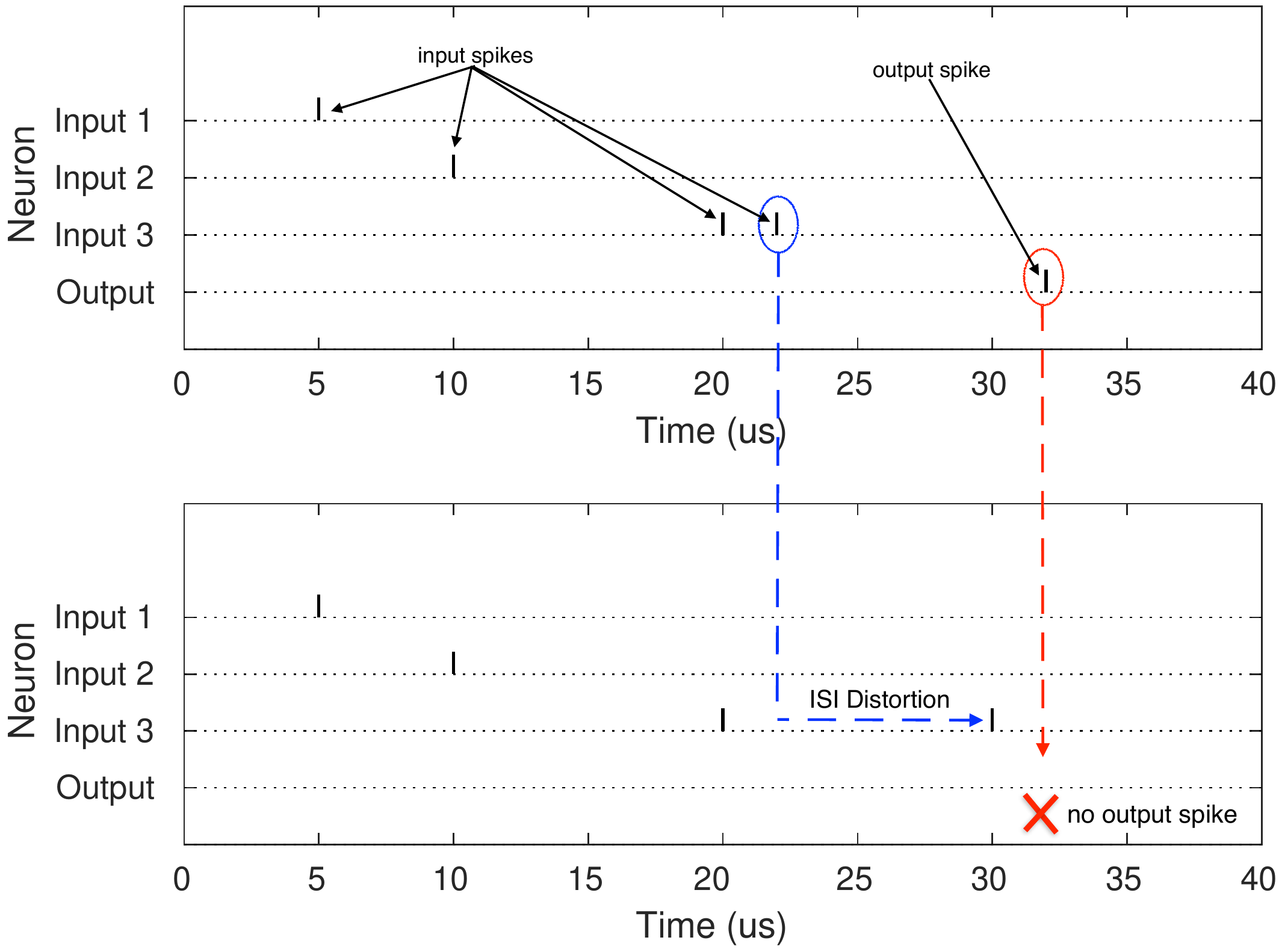}}
	\caption{Impact of ISI distortion on accuracy~\cite{pycarl}. Top sub-figure shows a scenario where an output spike is generated based on the spikes received from the three input neurons. Bottom sub-figure shows a scenario where the second spike from neuron 3 is delayed. There are no output spikes  generated.}
	\label{fig:isi_imact}
\end{figure}

To illustrate how a change in ISI, called \textbf{ISI distortion}, impacts inference quality, we use a small SNN in which three input neurons are connected to an output neuron. Figure~\ref{fig:isi_imact} illustrates the impact of ISI distortion on the output spike. In the top sub-figure, a spike is generated at the output neuron at 22$\mu$s due to spikes from the input neurons. In the bottom sub-figure, the second spike from input 3 is delayed, i.e., it has an ISI distortion. Due to this distortion, there is no output spike generated. Missing spikes can impact inference quality, as spikes encode information in SNNs.

\begin{figure}[h!]%
    \centering
    \subfloat[Original Image.\label{fig:original_image}]{{\includegraphics[width=4.0cm]{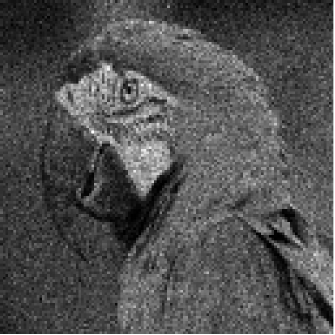} }}%
    \hfill
    \subfloat[Output with no ISI distortion (PSNR = 20).\label{fig:0ms_image}]{{\includegraphics[width=4.0cm]{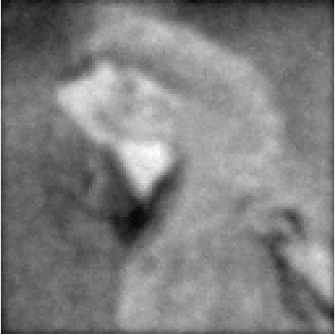}}}%
    \hfill
    \subfloat[Output with ISI distortion (PSNR = 19).\label{fig:20ms_image}]{{\includegraphics[width=4.0cm]{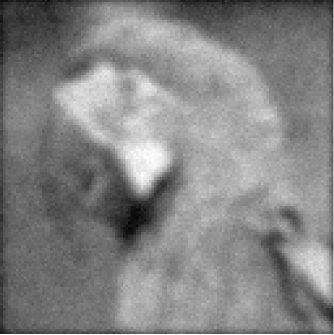}}}%
    \caption{Impact of ISI distortion on image smoothing.}%
    \label{fig:ISI_impact}%
\end{figure}

Figure~\ref{fig:ISI_impact} shows the impact of ISI distortion on the quality of image smoothing implemented using an SNN~\cite{carlsim}.
Figure~\ref{fig:original_image} shows the input image, which is fed to the SNN. Figure~\ref{fig:0ms_image} shows the output of the image smoothing application with no ISI distortion. PSNR of the output with reference to the input is 20. Figure~\ref{fig:20ms_image} shows the output with ISI distortion due to variation in latency within neuromorphic PEs of the hardware. PSNR of this output with respect to the input is 19. A reduction in PSNR indicates that the output image quality with ISI distortion is lower than the one without distortion. In fact, image quality deteriorates with increase in ISI distortion. \mr{We use ISI distortion as a measure of the quality of machine learning inference~\cite{spinemap}.} Our aim is to improve this inference quality via technological and architectural enhancements that reduce ISI distortion when the inference task is implemented on neuromorphic PEs of a hardware.

\section{Hardware Implementation of Machine Learning Inference}
\begin{figure}[h!]
	\centering
	\centerline{\includegraphics[width=0.99\columnwidth]{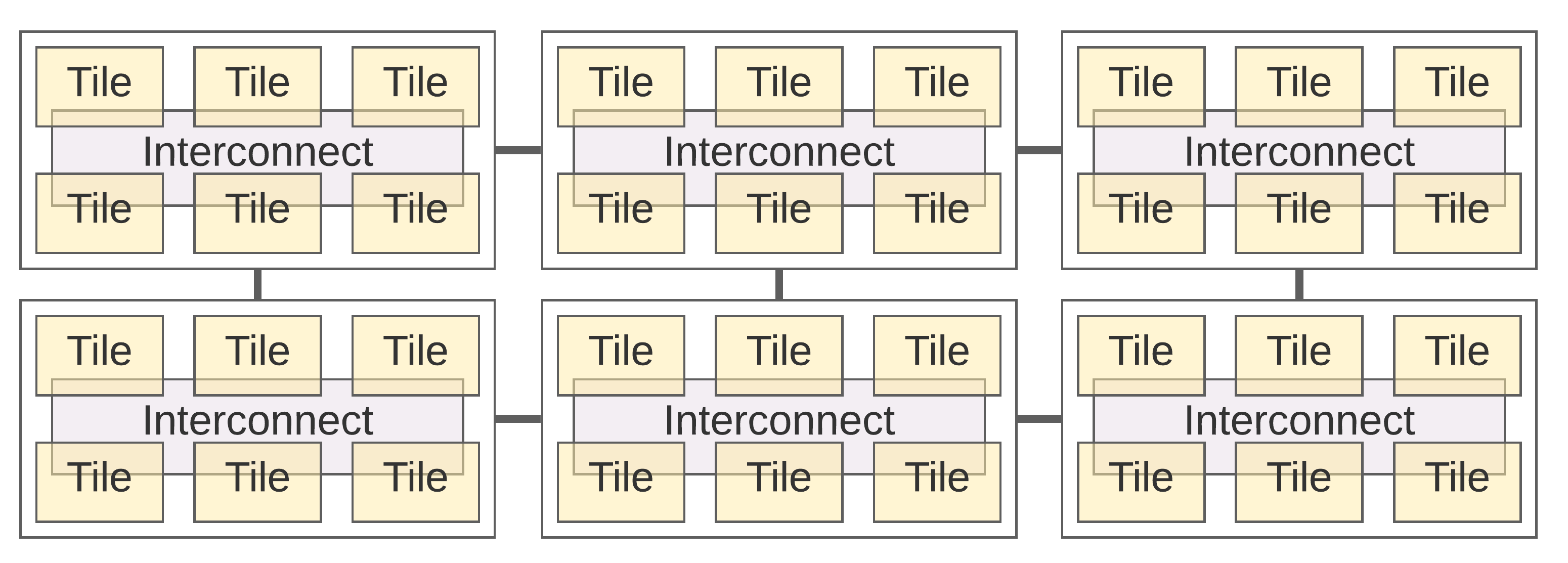}}
	\caption{Tile-based neuromorphic hardware, representative of hardware platforms such as TrueNorth~\cite{truenorth}, Loihi~\cite{loihi}, DYNAPs~\cite{dynapse}, and \mubrain{}~\cite{sentryos}.}
	\label{fig:tile}
\end{figure}

Most neuromorphic hardware platforms are implemented as tiled-based architectures~\cite{rajendran2019low,frenkel2020bottom,catthoor2018very,truenorth,loihi,sentryos}, where the tiles are interconnected via a shared interconnect such as Network-on-Chip~\cite{liu2018neu} and Segmented Bus~\cite{balaji2019exploration}.
Figure~\ref{fig:tile} illustrates a tile-based neuromorphic hardware platform, where the tiles can communicate concurrently. 
\mr{
Each tile includes 1) a neuromorphic PE, which consists of neuron and synapse \ckts{} and 2) a network interface, which encodes spikes into Address Event Representation (AER) and communicates these AER packets to the switch for routing to their destination tiles. A common design practice is to use analog crossbars to implement a neuromorphic PE~\cite{liu2015spiking,hu2014memristor,zhang2018neuromorphic,ankit2017trannsformer,kim2015reconfigurable,li2021hardware,spinemap,le2021memristor}.
}
Within a crossbar, a pre-synaptic neuron circuit acts as a current driver and is placed on a wordline, while a post-synaptic neuron circuit acts as a current sink and is placed on a bitline as illustrated in Figure~\ref{fig:crossbar} (left).

\mr{
Since a crossbar can accommodate only a limited number of neurons and synapses, a machine learning model is first partitioned into clusters, where each cluster can be implemented on a crossbar of the hardware. Partitioned clusters are then mapped to different crossbars when admitting the model to the hardware platform. To this end, several heuristic approaches are proposed in literature. PSOPART~\cite{psopart} minimizes spike latency on the shared interconnect, SpiNeMap~\cite{spinemap} minimizes interconnect energy, DFSynthesizer~\cite{dfsynthesizer_pp} maximizes throughput, DecomposedSNN~\cite{esl20} maximizes crossbar utilization, EaNC~\cite{twisha_energy} minimizes overall energy of a machine learning task by targeting both computation and communication energy, TaNC~\cite{twisha_thermal} minimizes the average temperature of each crossbar, eSpine~\cite{espine} maximizes NVM endurance in a crossbar, RENEU~\cite{reneu} minimizes the circuit aging in a crossbar's peripheral circuits, and NCil~\cite{song2021improving} reduces read disturb issues in a crossbar, improving the inference lifetime.
Beside these techniques, there are also other software frameworks~\cite{corelet,pacman,loihi_mapping,balaji2019framework,shihao_soda,shihao_designflow,neuroxplorer,vts_das,shihao_igsc,adarsha_igsc,twisha_endurance,song2020case,dfsynthesizer,balaji2019ISVLSIframework,das2018dataflow,ji2016neutrams,paul2022mitigation,huynh2022implementing} and run-time approaches~\cite{ncrtm,balaji2020run}, addressing one or more of these optimization objectives.
}


\mr{
We investigate the internal architecture of a crossbar and find that the parasitic components introduce delay in propagating current from a pre-synaptic neuron to a  post-synaptic neuron as illustrated in Figure~\ref{fig:crossbar} (right). 
This delay depends on the specific current path used in the mapping.
}
Higher the number of parasitic components on a current path, larger is its propagation delay. Parasitic components on bitlines and wordlines are a major source of latency at scaled process technology nodes and they create significant \textbf{latency variation} in a crossbar. Specifically, the latency of a synaptic connection in an SNN depends precisely on the memory cell in the crossbar that is used to implement it. Such latency variation can introduce ISI distortion (Section~\ref{sec:isi_distortion}), which may impact the quality of an inference task.

\section{Non-Volatile Memory Technology}
\begin{figure}[h!]
	\begin{center}
		\includegraphics[width=0.69\columnwidth]{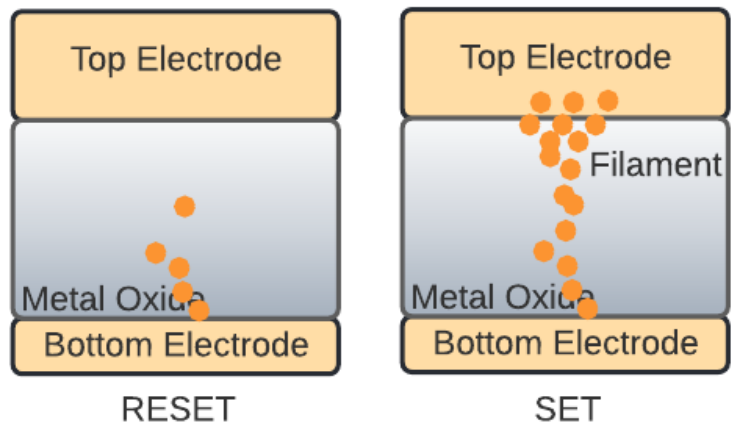}
		\caption{Operation of an RRAM cell with the $\text{HfO}_2$ layer sandwiched between the metals Ti (top electrode) and TiN (bottom electrode). The right subfigure shows the formation of LRS/SET state. The left subfigure shows the HRS/RESET state.}
		\label{fig:RRAM}
	\end{center}
\end{figure}

RRAM technology presents an attractive option for implementing memory cells of a crossbar due to its demonstrated potential for low-power multilevel operation and high integration density~\cite{mallik2017design}. An RRAM cell is composed of an insulating film sandwiched between conducting electrodes forming a metal-insulator-metal (MIM) structure (see Figure~\ref{fig:RRAM}). Recently, conducting filament-based metal-oxide RRAM implemented with transition-metal-oxides such as HfO${}_2$, ZrO${}_2$, and TiO${}_2$ has received considerable attention due to their low-power and CMOS-compatible scaling.

Synaptic weights are represented as conductance of the insulating layer within each RRAM cell. To program an RRAM cell, elevated voltages are applied at the top and bottom electrodes, which re-arranges the atomic structure of the insulating layer. Figure~\ref{fig:RRAM} shows the High-Resistance State (HRS) and the Low-Resistance State (LRS) of an RRAM cell. An RRAM cell can also be programmed into intermediate low-resistance states, allowing its multilevel operations~\cite{chen2015compact}.

\end{document}